\documentclass[conference]{IEEEtran}
\IEEEoverridecommandlockouts

\usepackage{cite}
\usepackage{amsmath,amssymb,amsfonts}
\usepackage{algorithmic}
\usepackage{graphicx}
\usepackage{textcomp}
\usepackage{xcolor}
\def\BibTeX{{\rm B\kern-.05em{\sc i\kern-.025em b}\kern-.08em
    T\kern-.1667em\lower.7ex\hbox{E}\kern-.125emX}}

\usepackage{subfig}
\usepackage{array,multirow,graphicx}
\usepackage{booktabs}
\usepackage{algorithmic}
\usepackage{bbm}
\usepackage{algorithm}
\usepackage{mhchem}

\begin{document}

\title{Explanation Space: \\ A New Perspective into Time Series Interpretability*\\
}

\author{\IEEEauthorblockN{1\textsuperscript{st} Shahbaz Rezaei}
\IEEEauthorblockA{\textit{Computer Science Department} \\
\textit{University of California}\\
Davis, CA, USA \\
srezaei@ucdavis.edu}
\and
\IEEEauthorblockN{2\textsuperscript{nd} Xin Liu}
\IEEEauthorblockA{\textit{Computer Science Department} \\
\textit{University of California}\\
Davis, CA, USA \\
xinliu@ucdavis.edu}
}

\maketitle

\begin{abstract}
Human understandable explanation of deep learning models is essential for various critical and sensitive applications. Unlike image or tabular data where the importance of each input feature (for the classifier's decision) can be directly projected into the input, time series distinguishable features (e.g. dominant frequency) are often hard to manifest in time domain for a user to easily understand. Additionally, most explanation methods require a baseline value as an indication of the absence of any feature. However, the notion of lack of feature, which is often defined as black pixels for vision tasks or zero/mean values for tabular data, is not well-defined in time series. Despite the adoption of explainable AI methods (XAI) from tabular and vision domain into time series domain, these differences limit the application of these XAI methods in practice. In this paper, we propose a simple yet effective method that allows a model originally trained on the time domain to be interpreted in other \textit{explanation spaces} using existing methods. We suggest five explanation spaces, each of which can potentially alleviate these issues in certain types of time series. Our method can be easily integrated into existing platforms without any changes to trained models or XAI methods. The code will be released upon acceptance.

\end{abstract}

\begin{IEEEkeywords}
XAI, uncertainty principle, time series.
\end{IEEEkeywords}

\section{Introduction}

Explainable AI (XAI) is a recent branch of studies dedicated to providing human interpretable explanation of deep models. Initially introduced for image data, attribution-based methods aim to highlight the region of the image responsible for the model's decision. Similar approaches have been later adopted for other data types, including time series. However, unlike computer vision, time series is an umbrella term covering vastly different applications with widely different characteristics and natures. This includes stock market data, EEG signal, audio signals, accelerometer data, as well as non-temporal data, such as spectrogram (e.g. Beef and Coffee dataset in UCR repository) or skeleton(contour)-based descriptors (e.g. ArrowHead or Fish datasets). Although application-blind DL approaches have often worked largely for classifications due to the power of DL, we face significant challenges when it comes to explanation. 

\begin{figure}[]
    \centering
    \subfloat[FordA (Time)]{%
        \includegraphics[width=0.45\linewidth, height=0.12\textwidth]{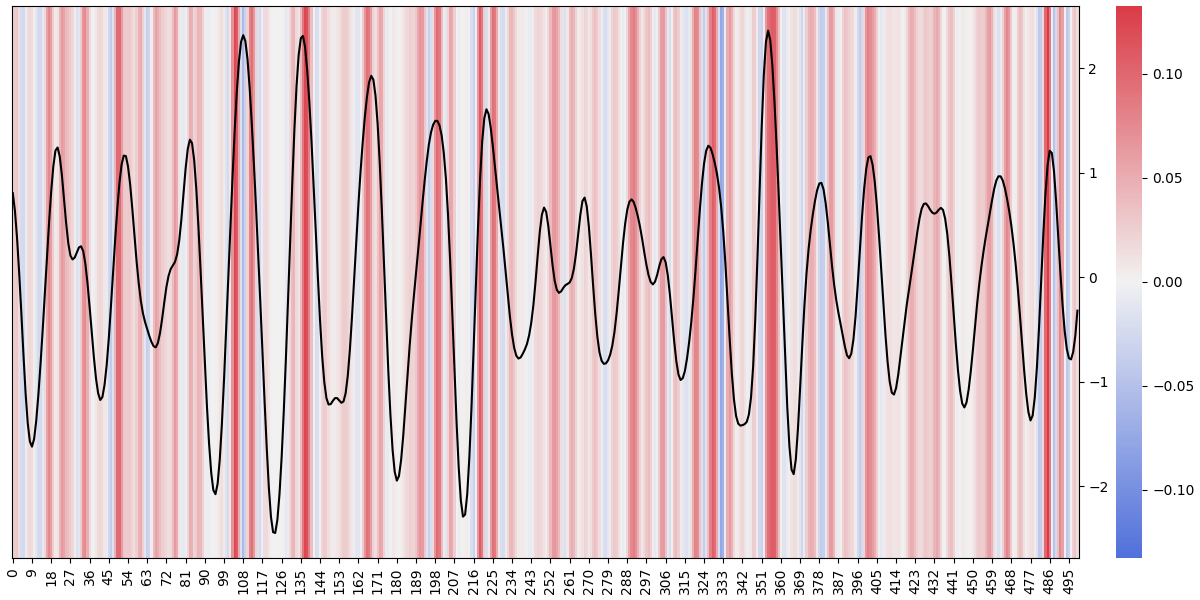}
        \label{fig:FrodA-time}}
    \subfloat[FordA (Freq.)]{
        \includegraphics[width=0.45\linewidth, height=0.12\textwidth]{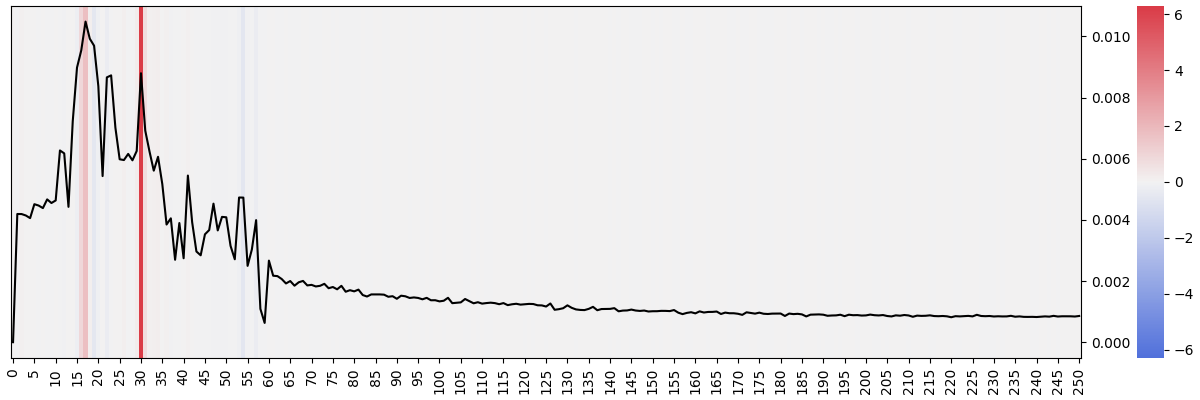}
        \label{fig:FrodA-freq}}
    \\
    \subfloat[AudioMNIST (Time)]{%
        \includegraphics[width=0.45\linewidth, height=0.12\textwidth]{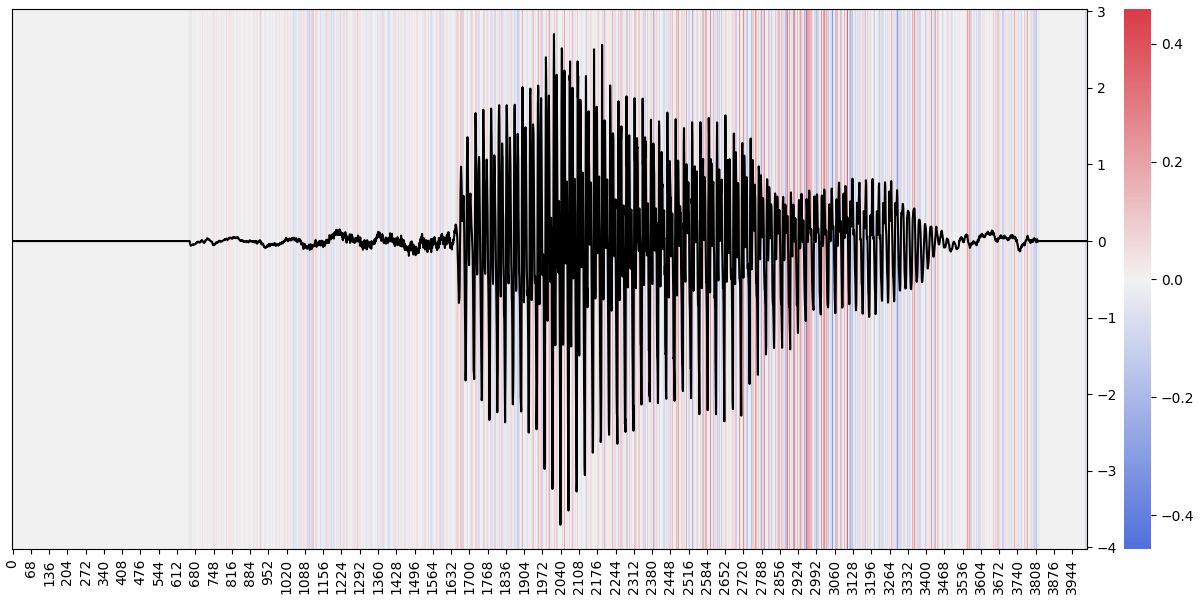}
        \label{fig:AudioMNIST-time}}
    \subfloat[AudioMNIST (Time/Freq)]{
        \includegraphics[width=0.45\linewidth, height=0.12\textwidth]{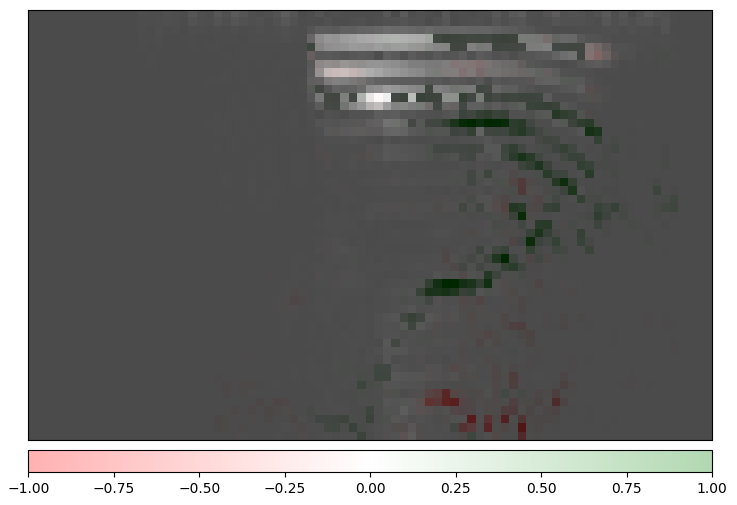}
        \label{fig:AudioMNIST-spec}}
    \caption{DeepLIFT attribution method on FordA and AudioMNIST datasets.}\label{fig:time-freq-example}
    \label{fig:FrodA-AudioMNIST}
\end{figure}

For instance, the explanation in time versus frequency domain of a model trained on FordA dataset is shown in Figure \ref{fig:FrodA-time} and \ref{fig:FrodA-freq} which clearly shows that only a few frequency components are dominant, and thus more desirable,  while the time domain explanation is more complex and hard to comprehend. Similarly, in audio signal the time/frequency space (Figure \ref{fig:AudioMNIST-spec}) generates a more sparse explanation than the time domain (Figure \ref{fig:AudioMNIST-time}).

In this paper, our premise is that a more  \textbf{sparse} explanation is more interpretable to an end user, and more likely to \textbf{reveal the nature} of the time series. Our key observation is that different spaces generate explanation with different sparsity on different type of time series that is aligned with our intuitive understanding of the nature of time series. More importantly, different spaces can reveal different aspects of data generation process, e.g. shapelet-based features in time domain and wave-based features in frequency. The main challenge is how to explain an existing DL model trained only on time domain in different domains.


To address this challenge, we introduce the concept of explanation space projection, where a simple procedure allows existing XAI methods to explain a model in different spaces\footnote{ In this paper, we use domain and space, interchangeably.}, such as time, frequency, or time/frequency. Using this procedure, we can generate explanations in multiple domains and automatically select the most appropriate space with respect to a chosen metric, such as sparsity. The main advantage of our method is that the target DL model is not required to be retrained on any other space. In other words, an existing model trained on any domain can be explained in all other domains using any existing XAI method. As a result, it can be easily adopted and used in practice.

We show that the explanation spaces are not limited to the well-known time and frequency spaces. In fact, any one-to-one function can introduce a new explanation space. To show the usefulness of this generalization in practice, we introduce three new explanation spaces that has not been used in XAI literature, namely, min-zero, difference, and decomposition spaces, each suitable for certain types of time-series. We evaluate our method on nine XAI methods using two DL models and different types of time series, including sensor data, audio, and motion, and illustrate the usefulness of different spaces in explaining different types of time series with less complexity.

\section{Related Work}

This paper focuses on post-hoc explanation methods.
There are several mechanisms to generate an explanation: Using gradient, such as Integrated Gradient \cite{sundararajan2017axiomatic}, GradientSHAP \cite{erion2019learning}, DeepLIFT \cite{shrikumar2017learning}, DeepSHAP \cite{lundberg2017unified}, Saliency \cite{simonyan2013deep}, InputXGradient \cite{shrikumar2016not}, Guided BackProp \cite{springenberg2014striving}, and KernelSHAP \cite{lundberg2017unified}, or occluding features, such as Feature Occlusion \cite{zeiler2014visualizing}, and feature ablation \cite{suresh2017clinical}. Most early works were originally developed for computer vision. Recently, some explanation methods are proposed to deal with challenges in time series data, such as TSR \cite{ismail2020benchmarking}, Demux \cite{doddaiah2022class}, TimeX \cite{queen2024encoding}, LEFTIST \cite{guilleme2019agnostic}, MILLET \cite{early2023inherently}, or DFT-LPR \cite{vielhaben2024explainable}. All these methods, and majority of computer vision methods, are attribution-based highlighting the importance of each pixel/time-step to the prediction.


In this paper, we focus on a new concept, i.e. explanation space, rather than a new explanation method. 
The closest work in literature is DFT-LPR \cite{vielhaben2024explainable} where the authors proposed a virtual inspection layer at the beginning of a model trained on time domain. They adopted LPR \cite{bach2015pixel} method such that it propagates the relevance back to the frequency or time/frequency domain. However, this method only works with LPR and also doesn't easily allow new spaces to be introduced. 
In this paper, we address this limitation by introducing a generalized wrapping mechanism that changes the input type of a model without changing its functionality. Hence, any existing XAI method can be used to generate explanation in the new input type. We show that the complexity introduced in the inspection layer of LPR is unnecessary and a simple, yet generalizable, approach can be adopted.

\section{Method}

Let $ x = \{x_1, x_2, ..., x_N\} \in R^{N}$ be a time series sample, where $x_i$ is the value at time step $i$ and $N$ is the number of time steps. To each time series sample there is a label associated, denoted by $y \in \{y_1, ..., y_C\}$ where $C$ is the number of classes. A classifier, $M : x \rightarrow y$, maps a times series sample to a probability distribution vector over $C$ classes. An explanation method, $E(M(\cdot), x) \rightarrow \bar{x} $, takes a classifier and a time series sample and generates an output, $\bar{x}$. In the case of attribution methods, $\bar{x}$ is a heat-map with the same dimension as $x$, and in the case of counterfactual methods, $\bar{x}$ is a new time series where $M(x) \neq M(\bar{x})$.

\textbf{Reversible Representation Space:}  A one-to-one function $F$, with a reverse function $F^{-1}$, is a representation space and it implies that $F^{-1}(F(x))=x$. In this paper, we also refer to it as an \textbf{explanation space} since we mainly focus on changing a domain for the purpose of explanation.

\textbf{Projection of Explanation from Time Domain into a New Explanation Space: } Assume a classifier $M$ trained on time domain. Consider an explanation space $F$ (e.g. FFT where it outputs frequency domain). Since $M(F^{-1}(F(x))) = M(x)$, we can define $z=F(x)$ as a new domain and the associated classifier is $M'(z) = M(F^{-1}(z)) \rightarrow y$. The new classifier $M'$ is the same as $M$ except that it performs the reverse operation, $F^{-1}$, before passing the input to $M$. Hence, $M$ on $x$ domain is functionally the same as $M'$ on $z$. However, now, we can use any existing explanation method, $E$, to generate an explanation on $z$ domain (or on explanation space $F$) using $E(M'(\cdot), z)$\footnote{It is desirable for $F/F^{-1}$ to be almost everywhere differentiable. Otherwise, gradient-based explanation methods may fail.}.

\subsection{Explanation Spaces}
The projection procedure allows us to generate an explanation on any explanation space even if the original model is only trained on time domain. Here, we introduce a few explanation spaces useful in certain applications:

\textbf{Frequency Space} is the Fourier transform and the reverse function is the inverse Fourier transform. As explained in Introduction, FordA dataset is an example where the frequency domain explanation is less complex than time (Figure \ref{fig:FrodA-freq} versus \ref{fig:FrodA-time}.

\textbf{Time/Frequency Space} is the short-time discrete Fourier transform to allow the explanation on both time and frequency domains. Figure \ref{fig:AudioMNIST-time} shows the attribution generated by DeepLIFT on AudioMNIST \cite{audiomnist2023} in time domain and Figure \ref{fig:AudioMNIST-spec} shows the attribution generated by the same method on time/frequency domain. The attribution on time/frequency domain is more focused on the center of the time series where actual signal is presented.

\begin{figure}[]
    \centering
    \subfloat[ElectricDevices (Time)]{%
        \includegraphics[width=0.45\linewidth, height=0.12\textwidth]{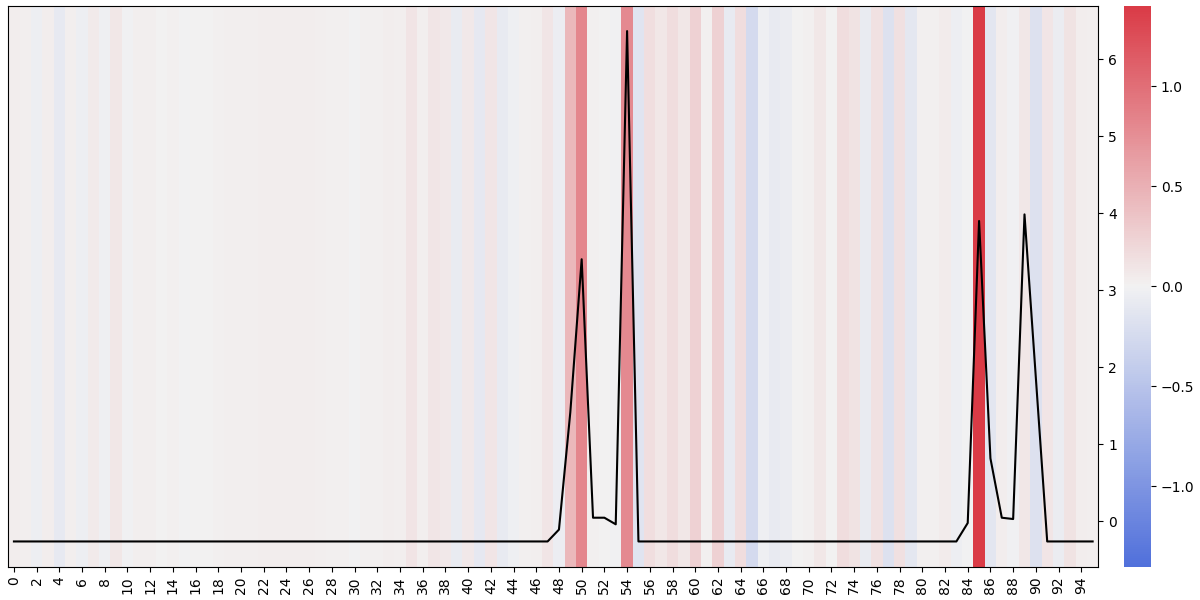}
        \label{fig:Elec-time}}
    \subfloat[ElectricDevices (Min zero)]{
        \includegraphics[width=0.45\linewidth, height=0.12\textwidth]{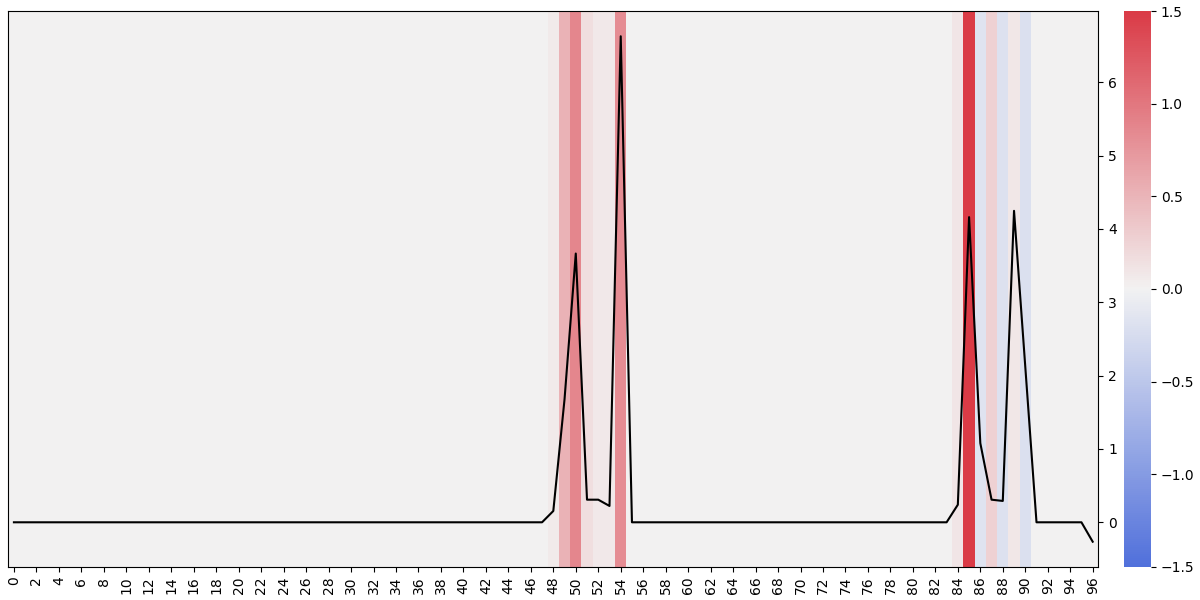}
        \label{fig:Elec-mz}}
    \\
    \subfloat[ECGFiveDays (Time)]{%
        \includegraphics[width=0.45\linewidth, height=0.12\textwidth]{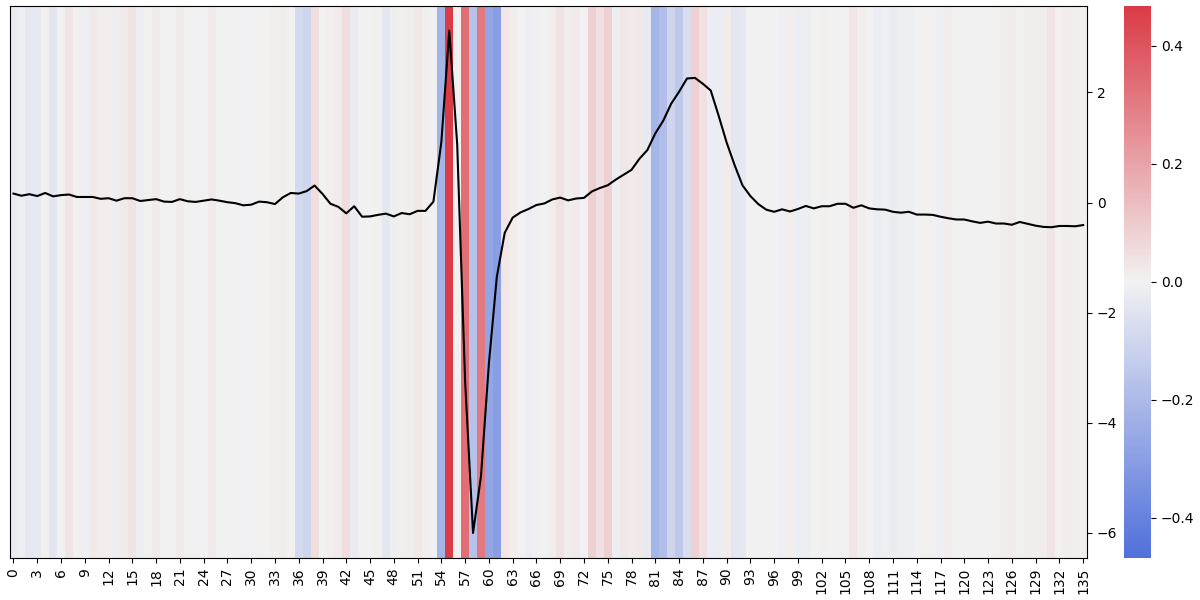}
        \label{fig:ECG-time}}
    \subfloat[ECGFiveDays (Diff.)]{
        \includegraphics[width=0.45\linewidth, height=0.12\textwidth]{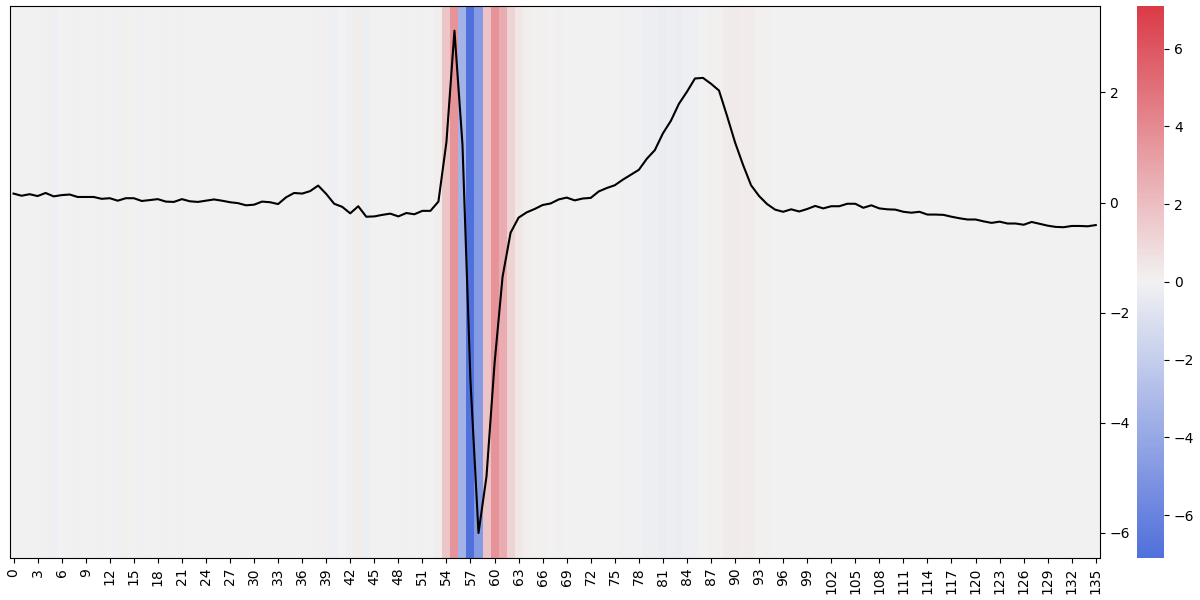}
        \label{fig:ECG-diff}}
    \\
    \subfloat[GunPoint (Time)]{%
        \includegraphics[width=0.45\linewidth, height=0.12\textwidth]{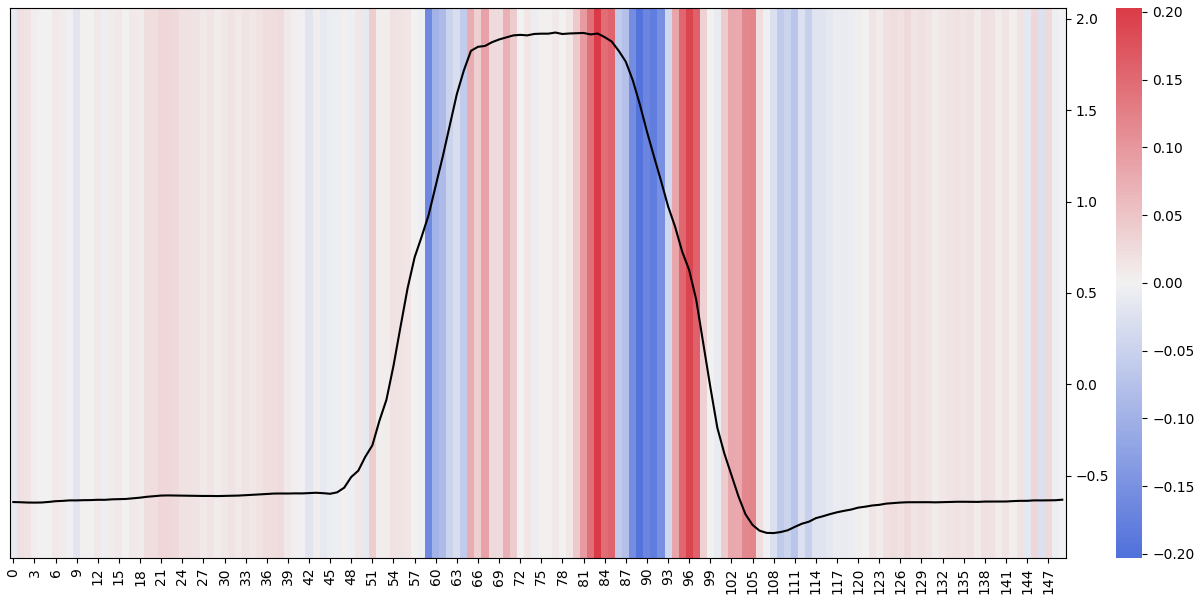}
        \label{fig:gun-time}}
    \subfloat[GunPoint (Freq.)]{
        \includegraphics[width=0.45\linewidth, height=0.12\textwidth]{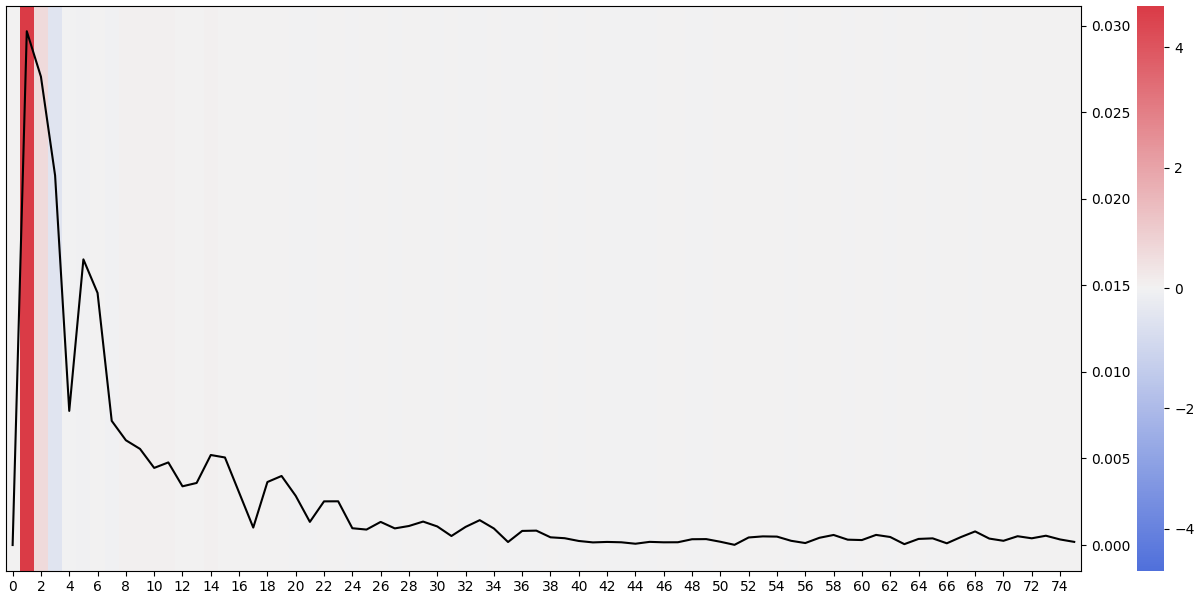}
        \label{fig:gun-freq}}
    \caption{DeepLIFT attribution method on ElectricDevices, ECGFiveDays, and GunPoint datasets.}
    \label{fig:mz-diff-example}
\end{figure}

\textbf{Min Zero Space} aims to alleviate the well-known issue of baseline exist on many attribution-based methods \cite{hollig2023xtsc}. By default, many existing explanation methods implicitly or explicitly assume a specific value (zero for the most part or a value set by user or training set) to be associated with a lack of any feature, including InputXGradient, feature occlusion, DeepLIFT, etc. Because time series are often z-normalized per sample to have a zero mean mainly for training stability, the zero value does not correspond to the lack of feature even if the original unnormalized time series does. For example, power consumption time series (e.g. ElectricDevices and LargeKitchenAppliances datasets in UCR repository) or time series dataset containing number of events or magnitude of events (e.g. Earthquakes and ChinaTown datasets in UCR repository or network traffic classification \cite{rezaei2020multitask}) are examples of natural time series where zero corresponds to a lack of feature/event in the unnormalized version, but not in the normalized one. 
This causes many artifacts to appear in XAI methods as shown in Figure \ref{fig:Elec-time}. This is consistent with the observation in \cite{samek2019explainable} that a global mean shift in data changes the explanation.

Min zero space maps the normalized time series on time domain to a min zero space domain where the minimum value is set back to zero such that it represents the lack of feature. To do so, $F_{min\_zero}$ shifts the entire time series such that the min is zero and then stores the previous min value in an extra time step at the end so that the reversible operation is possible, as follows:
\begin{multline}
    F_{min\_zero} (\{x_1, x_2, ..., x_N\}) \rightarrow \\
    \{x_1 - x_{min}, x_2 - x_{min}, ..., x_N -x_{min}, x_{min}\}
\end{multline}
where $x_{min}$ is the minimum value in the normalized time series. Note that the reverse operation ($F^{-1}$) is needed to be embedded into a new neural network, $M(F^{-1}(\cdot))$. This can be easily done by adding a single fully connected layer before $M()$ where it adds the last element to the first N elements. 
As shown in Figure \ref{fig:Elec-mz}, generating the attribution on mean zero space using the same method as Figure \ref{fig:Elec-time} completely removes all artifacts assigned to the part with no signal.

Note that, for some XAI methods, one can set minimum as a baseline value and generate an explanation in time domain similar to an explanation in min zero space, such as DeepLIFT or Occlusion. However, this is not true for all XAI methods, including Saliency which doesn't take a baseline, InputXGradient where the baseline is always implied to be zero, Integrated Gradient where the integral is taken over a range which may or may not include (start from) a target baseline, etc. In other words, the mean zero space provides a more general mechanism that allows all XAI methods to address the baseline issue for this type of TS.

\begin{figure}[]
    \centering
    \subfloat[Class 0 (Time)]{%
        \includegraphics[width=0.45\linewidth, height=0.12\textwidth]{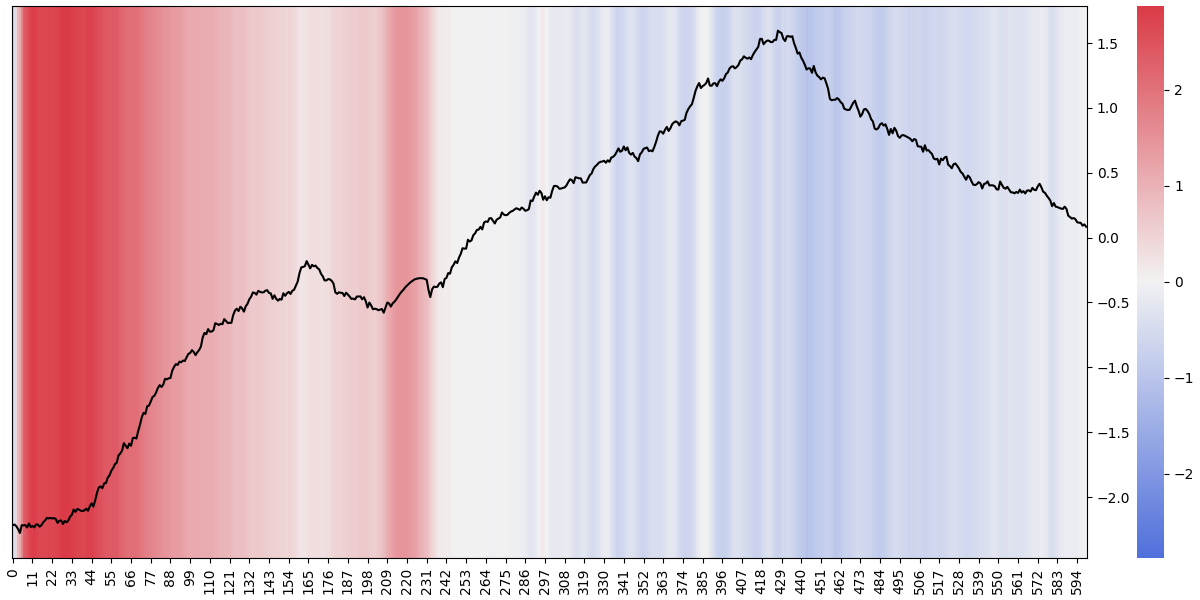}
        \label{fig:synth-time1}}
    \subfloat[Class 0 (Diff.)]{
        \includegraphics[width=0.45\linewidth, height=0.12\textwidth]{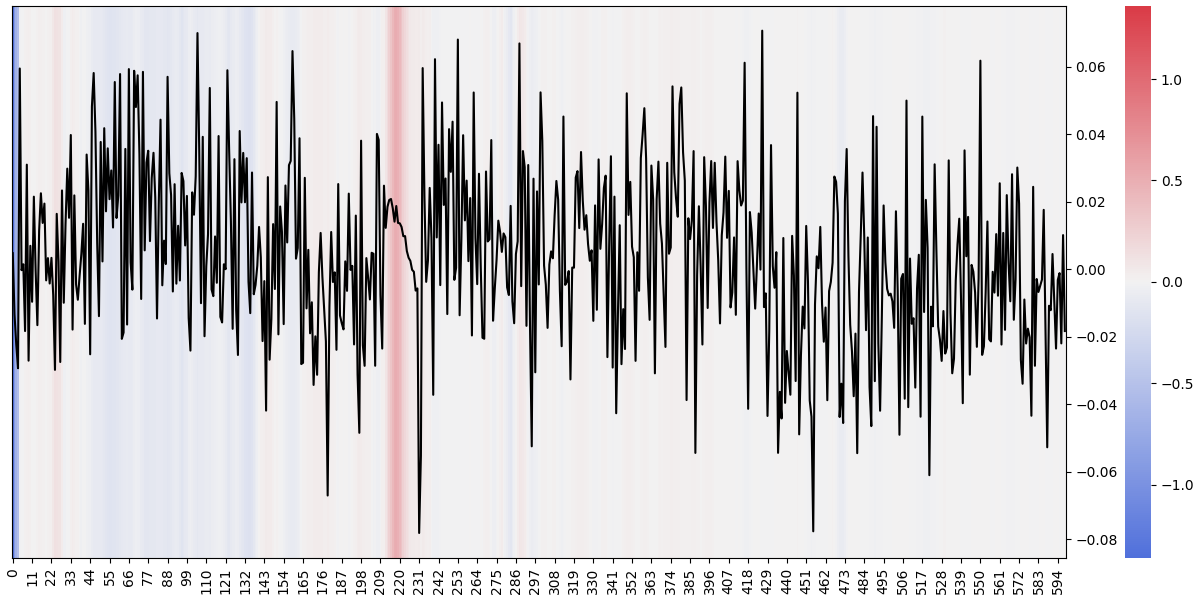}
        \label{fig:synth-diff1}}
    \caption{Occlusion XAI on a non-stationary dataset.}
    \label{fig:non-stationary-example}
\end{figure}

\textbf{Difference Space} is a result of taking a difference of each two consecutive time steps as a new time series. Differencing has been widely used in time series analysis to remove a trend or to make a time series stationary \cite{hyndman2018forecasting}. In our framework, difference domain has two benefits. First, for non-stationary time series (often with trend), such as stock  price or atmospheric time series, a shapelet-like feature in a small region can be completely imperceptible. For example, in a synthetic data generated to mimic a non-stationary time series with a small shapelet as a distinguishable feature, a shapelet feature may not be easily visible when it is small in comparison to the range of the entire time series\footnote{We use synthetic data in this experiment since there is no dataset with such properties in UCR repository.}, as shown in Figure \ref{fig:synth-time1}. However, the shapelets (i.e. semi-linear increase/decrease patterns visible in difference domain) can be clearly located in difference domain where the long range fluctuations is effectively removed, as shown in the corresponding samples in Figure \ref{fig:synth-diff1}. Interestingly, the difference and time domain has a huge impact on the performance of XAI methods. The XAI method completely fails the time domain which can be explained by the second benefit.

Second, the difference domain implicitly changes the notion of a feature. Here, the lack of change in time domain corresponds to zeros in difference domain (implying a lack of feature), and changes correspond to non-zeros values. This can potentially solve the baseline issue for certain types of time series.
For example, in time series associated with movements (e.g. GunPoint) or ECG signals, the lack of activity is manifested by a sequence of near constant values, not necessarily zeros. By taking the difference of the time series, a sequence of unchanging values is mapped to a sequence of zeros and, hence, aligns with the most explanation methods' baselines. Formally, the difference space is defined as
\begin{multline}
    F_{diff} (\{x_1, x_2, ..., x_N\}) \rightarrow \\
    \{x_1, x_2-x_1, x_3-x_2, ..., x_N-x_{N-1} \}
\end{multline}
The reverse operation is trivial.
Figure \ref{fig:ECG-time} and \ref{fig:ECG-diff} shows the attribution generated by DeepLIFT on time domain and difference domain, respectively. Note that in Figure \ref{fig:ECG-diff} we show the attribution generated on difference space on top of the time domain just for better visualization. In reality, the difference domain time series looks different from the original time domain. As shown, the difference space's explanation is sparser and focuses on the actual heartbeat of the ECG signal.

\begin{figure*}
\centering
\centering
\begin{tabular}{ccc}
\subfloat[Original TS]{\includegraphics[width=0.32\linewidth]{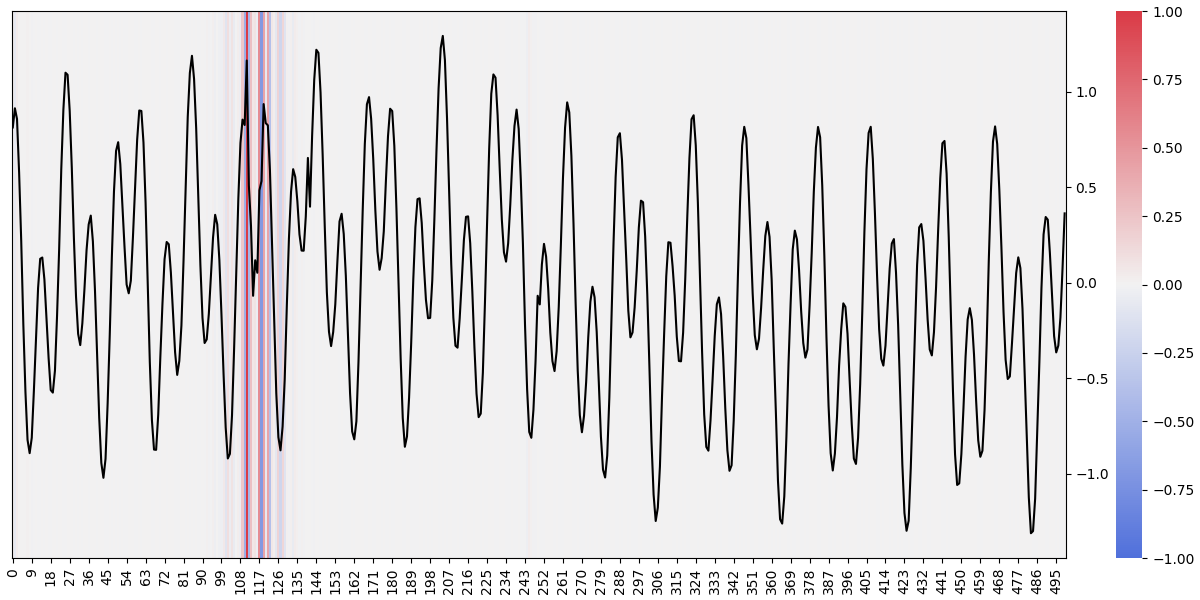} \label{fig:decomposition-original}} 
& \subfloat[Decomposed component 1]{\includegraphics[width=0.32\linewidth]{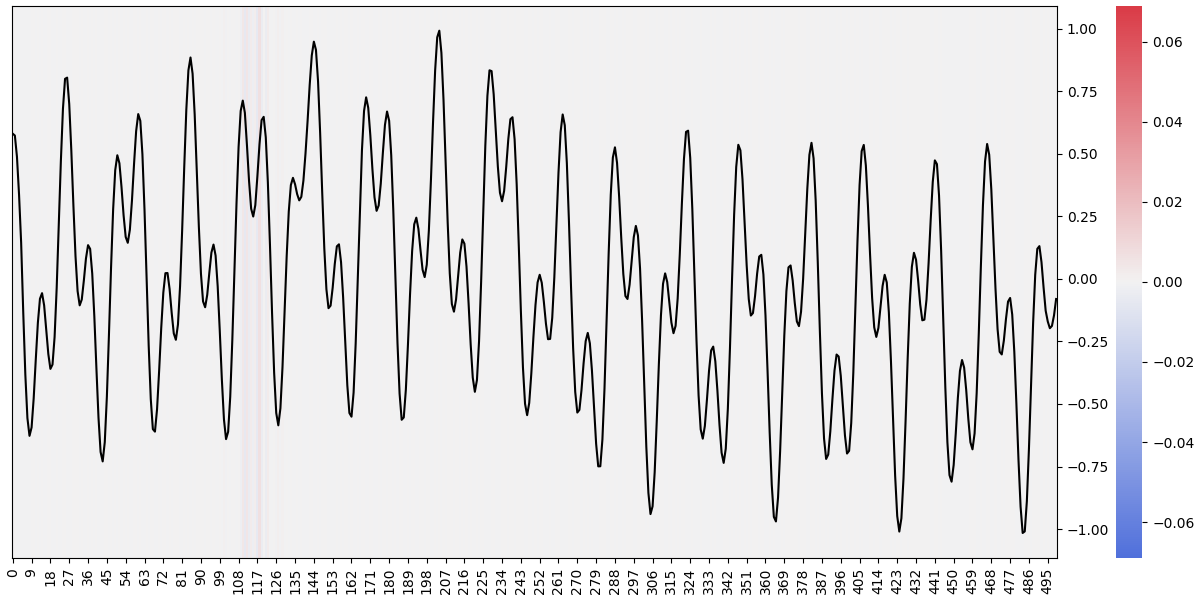}
\label{fig:decomposition-1}}
& \subfloat[Decomposed component 2]{\includegraphics[width=0.32\linewidth]{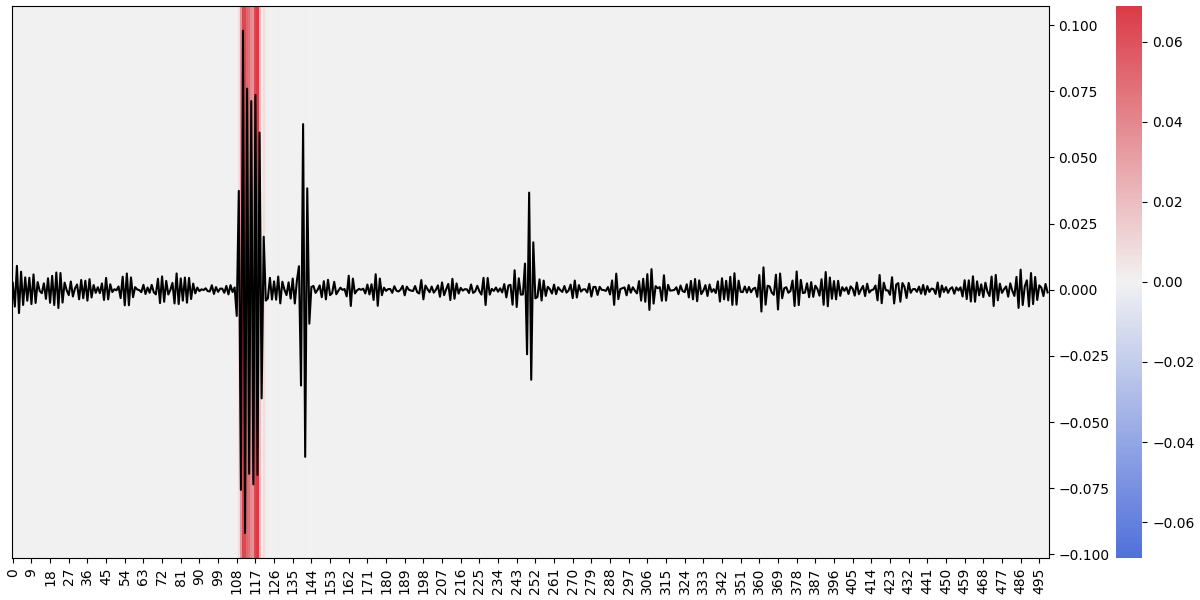}
\label{fig:decomposition-2}}\\
\end{tabular}
\caption{DeepLIFT attribution on a synthetic dataset for the original TS versus the calibrated attribution on 
 the decomposed components.}
\label{fig:decomposition-example}
\end{figure*}

\textbf{Decomposition Space} is a process of generating explanation on each component of the decomposed time series. Time series decomposition is a widely used method in many fields, such as financial time series \cite{guhathakurta2008empirical}, weather forecasting \cite{prema2015time}, traffic prediction \cite{huang2022short}, etc. Although deep models are mostly trained non-decomposed time series, attribution methods on the non-decomposed time series can be hard to interpret. For example, seasonality and trends can easily obscure the shapelet pattern. To solve this issue, we use the singular-spectrum analysis (SSA) decomposition method to get a set of time series with different components. Then, the reverse operation adds these components to reconstruct the original time series. Hence, we can generate an attribution for each of these components separately.
In Figure \ref{fig:decomposition-example}, we synthesize a dataset with different trends and multiple seasonality cycles plus some class-specific shapelet features. The ResNet model achieves $100\%$ accuracy when trained on the aggregated time series. As shown in Figure \ref{fig:decomposition-original}, DeepLIFT correctly highlights the location of the shapelet. However, because of the trend and seasonality, it is hard to identify the shapelet's pattern. The second decomposed component (Figure \ref{fig:decomposition-2}) containing the zigzag shapelet can be easily interpreted.

There is one caveat when using back-propagation methods for decomposition time series: Because the reverse operation  simply adds the decomposed components, the back-propagation methods essentially distribute the attribution similarly to all components. To solve this issue, we calibrate the attribution by combining the occlusion method with back-propagation methods as follows: Each decomposed component is occluded separately and the confidence drop is measured. Then, the confidence drop of each component is multiplied by the attribution values of that component. Finally, all attributions are normalized.

\section{Evaluation}

\textbf{Experimental Setup:} We include several datasets covering different types of time series, as shown in Table \ref{tbl-datasets-accuracy}, mainly from UCR Repository \cite{dau2019ucr} except AudioMNIST provided in \cite{vielhaben2024explainable}. For each dataset, we trained two commonly used model architectures: a ResNet\footnote{ResNet implementation for time series is adopted from TSEvo package \cite{hollig2022tsevo}.} and an InceptionTime model \cite{ismail2020inceptiontime}. We emphasize that all models are trained only on time domain. The explanations are generated on different domain using the process described in Method Section. We evaluate nine well-known XAI methods: DeepLIFT, GradientSHAP, Guided BackProp, InputXGradient (I$\times$G), Integrated Gradient (IG), KernelSHAP, LIME \cite{ribeiro2016should}, Occlusion, and Saliency \cite{simonyan2013deep}. We use Captum implementation \cite{kokhlikyan2020captum} for XAI methods. For min zero space, we ignore the attribution of the last time step, which is a placeholder for the min value. The reason is that perturbation of this single point, unlike other time points, changes the entire time series when the reverse operation is done. This causes the majority of XAI methods to give this point a very large attribution. Furthermore, none of the existing datasets in the UCR repository is suitable for the decomposition space since they are mostly short with no trends or seasonality. 

Note that the goal of this paper is not to have comprehensive evaluation of different XAI methods, as it has previously shown that no single XAI method can outperform others in all metrics/datasets \cite{loffler2022don}. Rather, we focus on illustrating that the concept of spaces is useful and that the same XAI method may bring a better explanation  on different spaces depending on the type of time series. 

\begin{table}
\scriptsize
  
  \centering
  \begin{tabular}{llll}
    \toprule
    Dataset & Type & ResNet & InceptionTime \\
    \midrule
    AudioMNIST & Audio & 87.47\% & 90.68\% \\
    FordA & Sensor & 93.56\% & 93.41\% \\
    GunPoint & Motion & 100\% & 99.33\% \\
    ECGFiveDays & ECG & 91.29\% & 100\% \\
    ECG5000 & ECG & 93.24\% & 94.47\% \\
    LargeKitchenAppliance & Device & 85.87\% & 88.00\% \\
    ElectricDevices & Device & 73.17\% & 71.22\% \\
    Earthquakes & Sensor/Event & 73.38\% & 74.82\% \\
    \bottomrule
  \end{tabular}
  \caption{Dataset types and models' accuracy}
  \label{tbl-datasets-accuracy}
\end{table}

\subsection{Evaluation Criteria}
\textbf{Robustness:} Robustness of an attribution method refers to its sensitivity with respect to small perturbations of the target sample \cite{hollig2023xtsc}. Unlike previous studies, we investigate the effect of different input spaces. As a result, a trained model wrapped to take frequency domain input may be more sensitive than the unwrapped version taking the time domain input. It is not trivial whether a model wrapped in other domains is more sensitive to perturbation than time domain or not. Hence, we also need to evaluate the sensitivity of the model itself in different input spaces. So, unlike previous work, we define two robustness: the first is associated with the classifier and the second is associated with the XAI method. \textit{Classifier robustness} is defined as follow:

\begin{equation}
    Rbt_{cls} = |M_c(x) - M_c(x + \lambda \epsilon)|
\end{equation}
where subscript $c$ indicates the probability output of the model for class $c$, the predicted class for sample $x$. Similar to \cite{hollig2023xtsc}, the XAI method's robustness is defined as follows:
\begin{equation}
    Rbt_{xai} = \frac{1}{|x|} || E(M(\cdot), x) - E(M(\cdot), x + \lambda \epsilon)||
\end{equation}

In our experiments, we take the average over 10 random perturbations. The perturbation, $\epsilon$, follows Gaussian distribution with zero mean and standard deviation equal to the input space's standard deviation, and $\lambda$ is set to $0.01$. 

\textbf{Faithfulness:} An explanation is faithful if the input features with highest attributions have the most effect on the prediction \cite{loffler2022don}. We measure the faithfulness using percentage of samples for which the prediction label flips when all non-negligible attributions are replaced with a default value, as follows:

\begin{equation}
    Faith = 1 - \mathbb{1}_{M(x)} \{ M(P(x, E, \epsilon)) \} 
\end{equation}
where $E$ is the attribution generated using sample $x$ and the model $M$, $\mathbb{1}_{M(x)}$ is the indicator function that return 1 when prediction class of $M(x)$ and $M(P(x, E, n))$ matches and zero otherwise, and $P(x, E, n)$ is the masking function that zero out input features of $x$ with attribution greater than $\epsilon$. Although the metric returns a boolean for a single sample, when taking the average over entire dataset, it shows the percentage of successful flip as a measure of faithfulness.


In the literature, some studies proposed using values rather than zero as a default baseline, either for generating explanation or measuring faithfulness. For example, \cite{loffler2022don} replaces a time step with the average value of that time step in training set (which is not a practical assumption for shift-variant time series), \cite{schlegel2019towards} picks a subsequence and reverses the time orders of the time steps, and \cite{parvatharaju2021learning} learns the most useful replacement/perturbation from the training data, etc. Such assumptions are always data type dependent. We avoid the baseline problem in all non-time domain spaces by defining them in such a way that zero has a specific meaning corresponding to certain lack of features. However, whether a certain space is suitable for a target dataset needs to be carefully considered by a practitioner familiar with that type of time series.

\begin{figure}[]
    \centering
    \subfloat[Time]{%
        \includegraphics[width=0.45\linewidth, height=0.12\textwidth]{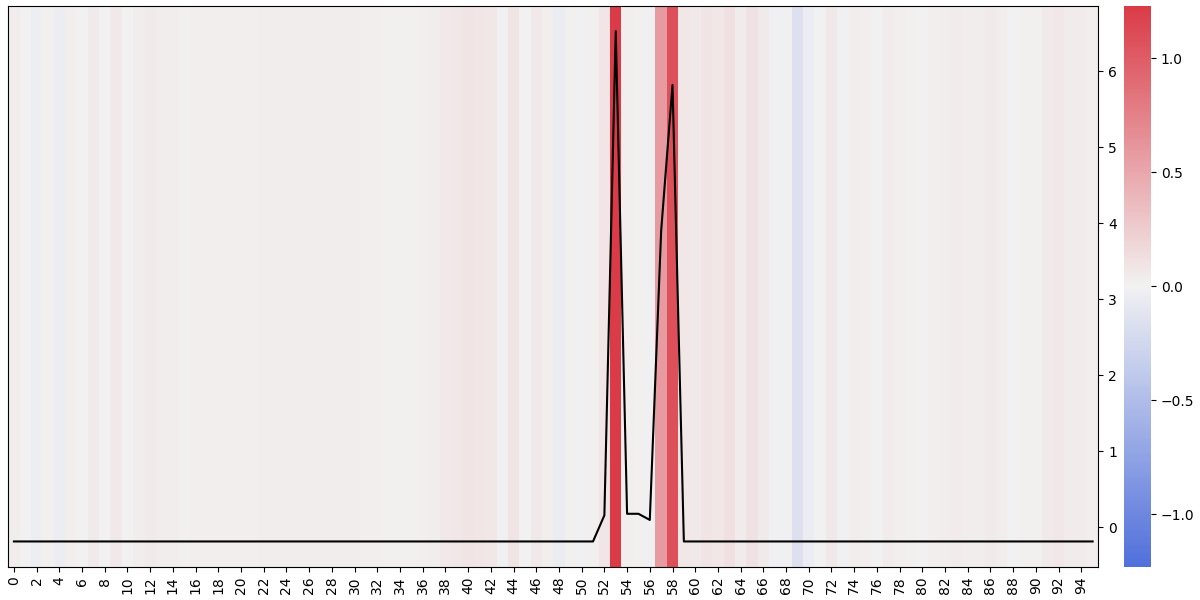}
        \label{fig:loc-time}}
    \subfloat[Frequency]{
        \includegraphics[width=0.45\linewidth, height=0.12\textwidth]{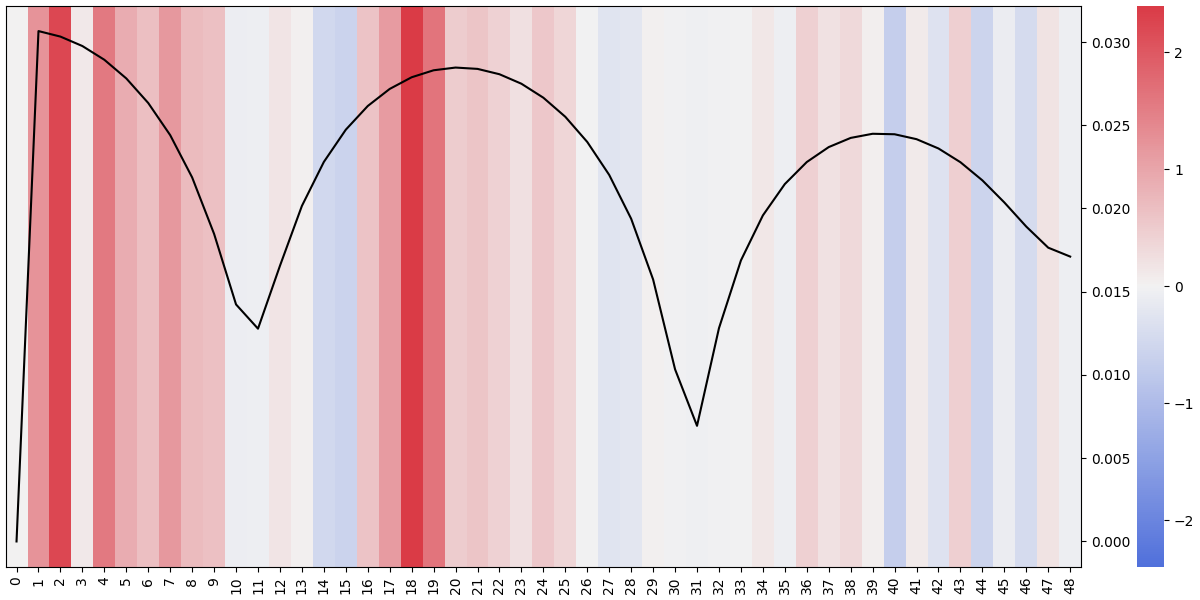}
        \label{fig:loc-freq}}
    \caption{DeepLIFT on ElectricDevices dataset. While the Shannon entropy indicates less complexity for the frequency domain (3.35 versus 3.50), our sparsity metric indicates less complexity for time domain (0.80 versus 0.40).}\label{fig:loc-example}
\end{figure}

\begin{table*}[h]
\scriptsize
  \centering
  \begin{tabular}{lllllllllll}
    \toprule
    Dataset & Domain & DeepLIFT & GradientSHAP & G. Backprop & I $\times$ G & IG & Kernel SHAP & LIME & Occlusion & Saliency\\
    \midrule
    \parbox[t]{1mm}{\multirow{5}{*}{\rotatebox[origin=c]{90}{AudioMNIST}}} & Time & 88\% (0.87) & 88\% (0.83) & 75\% (0.75) & 88\% (0.88) & 88\% (0.87) & 87\% (0.78) & 0\% (-) & 94\% (0.67) & 81\% (0.81) \\
    & Freq. & 97\% (0.87) & 94\% (0.89) & 99\% (\textbf{0.91}) & 95\% (0.86) & 96\% (0.86) & 75\% (\textbf{0.94}) & 34\% (-) & 97\% (0.62) & 99\% (0.80) \\
    & Time/Freq. & 100\% (\textbf{0.93}) & 99\% (\textbf{0.91}) & 99\% (0.84) & 100\% (\textbf{0.93}) & 75\% (\textbf{0.93}) & 91\% (0.82) & 15\% (-) & 100\% (\textbf{0.94}) & 92\% (\textbf{0.84}) \\
    & Difference & 90\% (0.88) & 90\% (0.81) & 100\% (0.07) & 90\% (0.88) & 90\% (0.88) & 90\% (0.87) & 45\% (-) & 93\% (0.75) & 100\% (0.21) \\
    & Min Zero & 90\% (0.75) & 90\% (0.71) & 98\% (0.75) & 89\% (0.82) & 91\% (0.76) & 98\% (0.66) & 2\% (-) & 81\% (0.94) & 88\% (0.81) \\
    
    \midrule
    \parbox[t]{1mm}{\multirow{5}{*}{\rotatebox[origin=c]{90}{FordA}}} & Time & 52\% (0.58) & 51\% (0.53) & 52\% (0.39) & 51\% (0.59) & 51\% (0.58) & 51\% (0.39) & 6\% (-) & 57\% (0.24) & 51\% (0.46) \\
    & Freq. & 97\% (\textbf{0.92}) & 61\% (\textbf{0.89}) & 94\% (\textbf{0.92}) & 90\% (\textbf{0.92}) & 98\% (\textbf{0.90}) & 63\% (0.67) & 14\% (-) & 85\% (0.88) & 82\% (\textbf{0.89}) \\
    & Time/Freq. & 100\% (0.87) & 84\% (0.84) & 75\% (0.78) & 100\% (0.87) & 90\% (0.83) & 82\% (0.58) & 20\% (-) & 96\% (\textbf{0.91}) & 97\% (0.66) \\
    & Difference & 51\% (0.43) & 51\% (0.42) & 51\% (0.10) & 51\% (0.45) & 51\% (0.43) & 51\% (\textbf{0.73}) & 49\% (-) & 52\% (0.34) & 51\% (0.11) \\
    & Min Zero & 51\% (0.44) & 51\% (0.54) & 51\% (0.39) & 51\% (0.46) & 51\% (0.40) & 51\% (0.40) & 26\% (-) & 52\% (0.87) & 51\% (0.46) \\
    
    \midrule
    \parbox[t]{1mm}{\multirow{4}{*}{\rotatebox[origin=c]{90}{GunPoint}}} & Time & 50\% (0.44) & 50\% (0.52) & 52\% (0.24) & 50\% (0.44) & 50\% (0.48) & 58\% (0.46) & 12\% (-) & 74\% (0.20) & 60\% (0.44) \\
    & Freq. & 68\% (\textbf{0.92}) & 54\% (\textbf{0.83}) & 86\% (\textbf{0.92}) & 54\% (\textbf{0.88}) & 64\% (\textbf{0.88}) & 64\% (\textbf{0.49}) & 38\% (-) & 82\% (\textbf{0.79}) & 96\% (\textbf{0.81}) \\
    & Difference & 64\% (0.65) & 62\% (0.55) & 66\% (0.11) & 66\% (0.64) & 70\% (0.69) & 58\% (0.44) & 4\% (-) & 86\% (0.51) & 76\% (0.13) \\
    & Min Zero & 48\% (-) & 64\% (0.56) & 52\% (0.24) & 66\% (0.60) & 54\% (0.60) & 68\% (0.44) & 22\% (-) & 66\% (0.20) & 70\% (0.44) \\

    \midrule
    \parbox[t]{1mm}{\multirow{4}{*}{\rotatebox[origin=c]{90}{ECG5Days}}} & Time & 39\% (-) & 43\% (-) & 43\% (-) & 39\% (-) & 39\% (-) & 39\% (-) & 26\% (-) & 61\% (0.43) & 39\% (-) \\
    & Freq. & 61\% (0.67) & 74\% (0.58) & 100\% (\textbf{0.77}) & 70\% (0.68) & 83\% (0.63) & 78\% (\textbf{0.55}) & 17\% (-) & 87\% (0.40) & 100\% (\textbf{0.61}) \\
    & Difference & 78\% (\textbf{0.81}) & 65\% (\textbf{0.80}) & 26\% (-) & 74\% (\textbf{0.80}) & 96\% (\textbf{0.82}) & 83\% (0.48) & 35\% (-) & 100\% (\textbf{0.61}) & 87\% (0.19) \\
    & Min Zero & 96\% (0.40) & 83\% (0.55) & 100\% (0.40) & 83\% (0.44) & 96\% (0.42) & 52\% (0.40) & 39\% (-) & 61\% (0.07) & 100\% (0.43) \\
    
    \midrule
    \parbox[t]{1mm}{\multirow{4}{*}{\rotatebox[origin=c]{90}{ECG5000}}} & Time & 4\% (-) & 4\% (-) & 3\% (-) & 5\% (-) & 5\% (-) & 3\% (-) & 1\% (-) & 95\% (0.28) & 85\% (0.44) \\
    & Freq. & 44\% (-) & 25\% (-) & 82\% (\textbf{0.81}) & 19\% (-) & 54\% (0.76) & 35\% (-) & 8\% (-) & 75\% (0.50) & 89\% (\textbf{0.65}) \\
    & Difference & 100\% (\textbf{0.83}) & 99\% (\textbf{0.76}) & 70\% (0.11) & 100\% (\textbf{0.82}) & 100\% (\textbf{0.80}) & 98\% (\textbf{0.50}) & 74\% (\textbf{0.81}) & 100\% (\textbf{0.75}) & 99\% (0.17) \\
    & Min Zero & 100\% (0.37) & 98\% (0.45) & 100\% (0.40) & 100\% (0.46) & 100\% (0.37) & 99\% (0.33) & 75\% (\textbf{0.81}) & 100\% (0.45) & 97\% (0.44) \\
    
    \midrule
    \parbox[t]{1mm}{\multirow{5}{*}{\rotatebox[origin=c]{90}{LargeKit.App.}}} & Time & 61\% (0.94) & 67\% (0.90) & 62\% (0.87) & 64\% (0.93) & 52\% (0.94) & 58\% (0.40) & 3\% (-) & 82\% (0.49) & 94\% (0.68) \\
    & Freq. & 88\% (0.78) & 61\% (0.77) & 84\% (0.75) & 79\% (0.84) & 60\% (0.80) & 57\% (\textbf{0.72}) & 38\% (-) & 97\% (0.65) & 87\% (\textbf{0.80}) \\
    & Time/Freq. & 99\% (\textbf{0.96}) & 68\% (0.90) & 38\% (-) & 94\% (0.94) & 100\% (0.95) & 49\% (-) & 30\% (-) & 67\% (0.90) & 82\% (0.79) \\
    & Difference & 84\% (\textbf{0.96}) & 85\% (\textbf{0.92}) & 82\% (0.16) & 84\% (\textbf{0.96}) & 84\% (\textbf{0.96}) & 92\% (0.51) & 49\% (-) & 99\% (\textbf{0.94}) & 100\% (0.07) \\
    & Min Zero & 65\% (\textbf{0.96}) & 52\% (\textbf{0.92}) & 61\% (\textbf{0.87}) & 64\% (\textbf{0.96}) & 51\% (\textbf{0.96}) & 72\% (0.69) & 3\% (-) & 69\% (0.89) & 80\% (0.68) \\
    
    \midrule
    \parbox[t]{1mm}{\multirow{4}{*}{\rotatebox[origin=c]{90}{ElectricDev.}}} & Time & 100\% (0.67) & 100\% (0.62) & 79\% (\textbf{0.50}) & 99\% (0.68) & 100\% (0.67) & 100\% (0.32) & 70\% (\textbf{0.76}) & 100\% (0.24) & 100\% (0.44) \\
    & Freq. & 98\% (0.60) & 92\% (0.51) & 100\% (0.44) & 95\% (0.54) & 98\% (0.54) & 88\% (0.42) & 80\% (0.74) & 99\% (0.27) & 99\% (\textbf{0.48}) \\
    & Difference & 99\% (0.65) & 100\% (0.64) & 85\% (0.11) & 99\% (0.64) & 99\% (0.64) & 99\% (\textbf{0.45}) & 79\% (0.72) & 99\% (0.47) & 99\% (0.14) \\
    & Min Zero & 67\% (\textbf{0.77}) & 75\% (\textbf{0.67}) & 79\% (\textbf{0.50}) & 71\% (\textbf{0.78}) & 75\% (\textbf{0.77}) & 65\% (0.43) & 40\% (0.76) & 92\% (\textbf{0.60}) & 85\% (0.45) \\
    
    \midrule
    \parbox[t]{1mm}{\multirow{5}{*}{\rotatebox[origin=c]{90}{Earthquakes}}} & Time & 82\% (0.87) & 73\% (0.79) & 79\% (\textbf{0.81}) & 78\% (0.84) & 82\% (0.85) & 50\% (0.71) & 2\% (-) & 77\% (0.65) & 68\% (\textbf{0.70}) \\
    & Freq. & 100\% (0.62) & 97\% (0.61) & 99\% (0.56) & 100\% (0.60) & 100\% (0.58) & 87\% (0.57) & 1\% (-) & 100\% (0.51) & 100\% (0.54) \\
    & Time/Freq. & 100\% (0.75) & 58\% (0.75) & 36\% (-) & 100\% (0.72) & 100\% (0.77) & 40\% (-) & 0\% (-) & 65\% (\textbf{0.87}) & 98\% (0.57) \\
    & Difference & 90\% (0.67) & 90\% (0.56) & 94\% (0.09) & 90\% (0.68) & 90\% (0.68) & 90\% (\textbf{0.64}) & 86\% (\textbf{0.94}) & 92\% (0.51) & 95\% (0.14) \\
    & Min Zero & 79\% (\textbf{0.90}) & 62\% (\textbf{0.80}) & 80\% (\textbf{0.81}) & 70\% (\textbf{0.90}) & 77\% (\textbf{0.90}) & 41\% (-) & 2\% (-) & 74\% (0.70) & 66\% (\textbf{0.70}) \\
    \bottomrule
  \end{tabular}
  \caption{ResNet: The first number is the faithfulness and the number in the parenthesis is the sparsity metric.}
  \label{tbl-resnet}
\end{table*}

\begin{table*}[h]
\scriptsize
  \centering
  \begin{tabular}{lllllllllll}
    \toprule
    Dataset & Domain & DeepLIFT & GradientSHAP & G. Backprop & I $\times$ G & IG & Kernel SHAP & LIME & Occlusion & Saliency\\
    \midrule
    \parbox[t]{1mm}{\multirow{5}{*}{\rotatebox[origin=c]{90}{AudioMNIST}}} 
    & Time & 66\% (0.91) & 87\% (0.85) & 80\% (0.78) & 86\% (0.88) & 87\% (0.88) & 78\% (0.79) & 0\% (-) & 92\% (0.77) & 93\% (0.75) \\
    & Freq. & 83\% (0.93) & 89\% (0.91) & 97\% (\textbf{0.91}) & 92\% (0.89) & 90\% (0.88) & 68\% (0.89) & 21\% (-) & 92\% (0.72) & 98\% (\textbf{0.83}) \\
    & Time/Freq. & 65\% (\textbf{0.97}) & 97\% (\textbf{0.93}) & 99\% (0.88) & 95\% (\textbf{0.95}) & 85\% (\textbf{0.93}) & 89\% (0.78) & 18\% (-) & 100\% (\textbf{0.96}) & 90\% (\textbf{0.83}) \\
    & Difference & 91\% (0.86) & 100\% (0.82) & 80\% (0.12) & 100\% (0.86) & 100\% (0.87) & 100\% (\textbf{0.91}) & 62\% (\textbf{0.99}) & 100\% (0.72) & 100\% (0.07) \\
    & Min Zero & 88\% (0.71) & 100\% (0.67) & 97\% (0.78) & 99\% (0.75) & 100\% (0.70) & 97\% (0.70) & 34\% (-) & 89\% (0.91) & 96\% (0.75) \\
    
    \midrule
    \parbox[t]{1mm}{\multirow{5}{*}{\rotatebox[origin=c]{90}{FordA}}} 
    & Time & 0\% (-) & 42\% (-) & 42\% (0.29) & 43\% (-) & 42\% (-) & 43\% (-) & 7\% (-) & 47\% (-) & 44\% (-) \\
    & Freq. & 0\% (-) & 62\% (\textbf{0.88}) & 79\% (\textbf{0.92}) & 83\% (\textbf{0.91}) & 90\% (\textbf{0.89}) & 47\% (-) & 16\% (-) & 74\% (0.88) & 51\% (\textbf{0.88}) \\
    & Time/Freq. & 0\% (-) & 60\% (0.86) & 62\% (0.80) & 99\% (0.89) & 84\% (0.85) & 66\% (\textbf{0.63}) & 15\% (nan) & 78\% (\textbf{0.92}) & 92\% (0.78) \\
    & Difference & 0\% (-) & 42\% (-) & 23\% (-) & 42\% (-) & 42\% (-) & 43\% (-) & 42\% (-) & 43\% (0.37) & 43\% (-) \\
    & Min Zero & 0\% (-) & 42\% (-) & 42\% (-) & 42\% (-) & 42\% (-) & 42\% (-) & 34\% (-) & 42\% (-) & 42\% (-) \\
    
    \midrule
    \parbox[t]{1mm}{\multirow{4}{*}{\rotatebox[origin=c]{90}{GunPoint}}} 
    & Time & 42\% (-) & 48\% (-) & 70\% (0.47) & 48\% (-) & 48\% (-) & 48\% (-) & 48\% (-) & 72\% (0.22) & 48\% (-) \\
    & Freq. & 30\% (-) & 52\% (\textbf{0.86}) & 96\% (\textbf{0.91}) & 50\% (\textbf{0.90}) & 96\% (\textbf{0.90}) & 72\% (0.47) & 50\% (\textbf{0.85}) & 94\% (\textbf{0.84}) & 94\% (\textbf{0.86}) \\
    & Difference & 72\% (\textbf{0.59}) & 70\% (0.52) & 48\% (0.24) & 70\% (0.57) & 70\% (0.67) & 68\% (\textbf{0.49}) & 60\% (0.82) & 70\% (0.51) & 70\% (0.08) \\
    & Min Zero & 48\% (-) & 48\% (-) & 48\% (-) & 48\% (-) & 48\% (-) & 48\% (-) & 44\% (-) & 48\% (-) & 48\% (-) \\
    
    \midrule
    \parbox[t]{1mm}{\multirow{4}{*}{\rotatebox[origin=c]{90}{ECGFiveDays}}} 
    & Time & 65\% (\textbf{0.83}) & 87\% (0.73) & 100\% (0.66) & 96\% (0.79) & 100\% (\textbf{0.78}) & 57\% (0.42) & 39\% (-) & 100\% (0.70) & 52\% (0.44) \\
    & Freq. & 65\% (0.70) & 83\% (0.73) & 96\% (\textbf{0.70}) & 100\% (0.72) & 100\% (0.74) & 87\% (0.38) & 26\% (-) & 91\% (0.72) & 100\% (\textbf{0.70}) \\
    & Difference & 61\% (\textbf{0.83}) & 65\% (\textbf{0.78}) & 70\% (0.18) & 70\% (\textbf{0.80}) & 35\% (-) & 91\% (\textbf{0.48}) & 96\% (\textbf{0.81}) & 100\% (\textbf{0.74}) & 83\% (0.21) \\
    & Min Zero & 61\% (0.67) & 83\% (0.65) & 100\% (0.66) & 100\% (0.55) & 100\% (0.56) & 61\% (0.40) & 52\% (0.80) & 100\% (0.26) & 87\% (0.44) \\
    
    \midrule
    \parbox[t]{1mm}{\multirow{4}{*}{\rotatebox[origin=c]{90}{ECG5000}}} 
    & Time & 68\% (0.66) & 81\% (0.59) & 96\% (0.52) & 95\% (0.61) & 89\% (0.69) & 75\% (0.52) & 9\% (-) & 99\% (0.59) & 100\% (0.38) \\
    & Freq. & 43\% (-) & 69\% (\textbf{0.74}) & 99\% (\textbf{0.83}) & 93\% (\textbf{0.80}) & 95\% (0.71) & 75\% (0.49) & 24\% (-) & 89\% (\textbf{0.72}) & 100\% (\textbf{0.74}) \\
    & Difference & 100\% (\textbf{0.81}) & 99\% (0.63) & 100\% (0.03) & 100\% (0.76) & 100\% (\textbf{0.77}) & 93\% (\textbf{0.53}) & 40\% (-) & 100\% (0.54) & 100\% (0.07) \\
    & Min Zero & 100\% (0.53) & 100\% (0.51) & 100\% (0.53) & 100\% (0.43) & 100\% (0.51) & 100\% (0.42) & 93\% (\textbf{0.81}) & 100\% (0.15) & 100\% (0.38) \\
    
    \midrule
    \parbox[t]{1mm}{\multirow{5}{*}{\rotatebox[origin=c]{90}{LargeKit.App.}}} 
    & Time & 43\% (-) & 66\% (\textbf{0.93}) & 66\% (\textbf{0.80}) & 70\% (0.94) & 65\% (0.95) & 73\% (\textbf{0.69}) & 11\% (-) & 94\% (0.86) & 87\% (0.70) \\
    & Freq. & 53\% (0.81) & 75\% (0.81) & 92\% (0.70) & 87\% (0.85) & 85\% (0.81) & 56\% (0.68) & 2\% (-) & 95\% (0.69) & 97\% (\textbf{0.82}) \\
    & Time/Freq. & 74\% (\textbf{0.97}) & 68\% (0.92) & 43\% (-) & 95\% (0.94) & 97\% (\textbf{0.96}) & 52\% (0.67) & 23\% (-) & 64\% (0.83) & 76\% (0.78) \\
    & Difference & 87\% (\textbf{0.97}) & 84\% (0.92) & 47\% (-) & 84\% (\textbf{0.96}) & 84\% (\textbf{0.96}) & 91\% (0.57) & 26\% (-) & 97\% (\textbf{0.91}) & 95\% (0.09) \\
    & Min Zero & 45\% (-) & 63\% (\textbf{0.93}) & 66\% (\textbf{0.80}) & 67\% (\textbf{0.96}) & 61\% (\textbf{0.96}) & 57\% (\textbf{0.69}) & 13\% (-) & 84\% (\textbf{0.91}) & 64\% (0.70) \\
    
    \midrule
    \parbox[t]{1mm}{\multirow{4}{*}{\rotatebox[origin=c]{90}{ElectricDev.}}} 
    & Time & 83\% (0.67) & 90\% (0.61) & 83\% (0.25) & 90\% (0.67) & 95\% (0.64) & 89\% (0.36) & 59\% (\textbf{0.77}) & 72\% (0.26) & 100\% (0.42) \\
    & Freq. & 95\% (0.74) & 91\% (0.52) & 99\% (\textbf{0.51}) & 85\% (0.51) & 97\% (0.53) & 92\% (\textbf{0.45}) & 78\% (0.76) & 97\% (0.28) & 99\% (\textbf{0.45}) \\
    & Difference & 99\% (0.64) & 68\% (0.63) & 46\% (-) & 92\% (0.66) & 88\% (0.65) & 98\% (\textbf{0.45}) & 79\% (0.73) & 93\% (0.49) & 96\% (0.11) \\
    & Min Zero & 81\% (\textbf{0.80}) & 81\% (\textbf{0.67}) & 83\% (0.25) & 79\% (\textbf{0.79}) & 89\% (\textbf{0.75}) & 70\% (0.43) & 51\% (0.75) & 82\% (\textbf{0.65}) & 88\% (0.42) \\
    
    \midrule
    \parbox[t]{1mm}{\multirow{5}{*}{\rotatebox[origin=c]{90}{Earthquakes}}} 
    & Time & 16\% (-) & 53\% (\textbf{0.69}) & 64\% (\textbf{0.48}) & 56\% (0.72) & 58\% (0.76) & 17\% (-) & 4\% (-) & 45\% (-) & 44\% (-) \\
    & Freq. & 17\% (-) & 47\% (-) & 18\% (-) & 93\% (0.65) & 95\% (0.64) & 17\% (-) & 3\% (-) & 82\% (\textbf{0.53}) & 40\% (-) \\
    & Time/Freq. & 79\% (\textbf{0.75}) & 46\% (-) & 22\% (-) & 86\% (0.73) & 83\% (0.72) & 28\% (-) & 5\% (-) & 52\% (0.47) & 86\% (\textbf{0.49}) \\
    & Difference & 15\% (-) & 15\% (-) & 16\% (-) & 15\% (-) & 15\% (-) & 15\% (-) & 15\% (-) & 16\% (-) & 30\% (-) \\
    & Min Zero & 16\% (-) & 42\% (-) & 90\% (\textbf{0.48}) & 51\% (\textbf{0.83}) & 51\% (\textbf{0.84}) & 18\% (-) & 5\% (0.95) & 41\% (-) & 44\% (-) \\
    \bottomrule
  \end{tabular}
    \caption{InceptionTime: he first number is the faithfulness and the number in the parenthesis is the sparsity metric.}
  \label{tbl-inceptiontime}
\end{table*}

\textbf{Sparsity: } In the literature, sparsity (also known as complexity) is measured using Shannon entropy \cite{hollig2023xtsc} \cite{vielhaben2024explainable}. Shannon entropy has a few limitations: First, it does not have a predefined upper limit. In other words, for a given dataset, Shannon entropy is always between zero and a positive real number that depends on the length of the time series. Because the upper limit is not fixed and predefined, it is hard for an XAI user to internalize how complex an explanation is in an absolute sense. Another problem with Shannon entropy is its \textit{length-dependent accumulation of extremely small non-zero attributions}. A motivating example of why the length matters is shown in Figure \ref{fig:loc-example}. Since the length of the input in time domain is twice the length in frequency domain and because the attribution in time domain contains many small non-zero values, all non-zero values accumulate resulting in the higher Shannon entropy in time domain. This is in stark contrast to what is visually presented. Our formulation (in equation (\ref{eq-loc})), however, indicates better sparsity since it takes the length into account.

Last, ignoring the length may have other consequences. For example, imagine two different input domains, $X_1$ and $X_2$, where the length of the first one is larger than the second one, i.e. $l_1 >> l_2$. If an XAI method assigns high attribution to two input features and zero to others in both input domains, Shannon entropy of the two explanations is equal. However, picking out two input features in an space with twice the length is more significant. Unlike previous studies, we need our metric to be sensitive to the input space length so that we can compare explanations in different spaces. We note that the choice of an explanation space and which metric is appropriate is not trivial and depends on the nature of the application. Hence, a practitioner should choose a suitable option accordingly.

Here, we introduce four desirable properties for a sparsity metric and then propose a new simple metric that satisfies all properties. A sparsity metric, $Spr(\cdot)$, should satisfy:
\begin{enumerate}
    \item Sparsity is bounded, i.e. $0 \leq Spr(E(x)) \leq 1$, where greater values mean more sparsity (i.e. less complexity).
    \item A complete uniform attribution has zero sparsity, i.e. if $E_i(x) = E_j(x)$ for all $1 \leq i,j \leq n$, then $Spr(E(x)) = 0$. Here, $E_i$ indicates the attribution at input feature $i$.
    \item An explanation with single non-zero attribution at $i$ and zero elsewhere has a sparsity value of 1, i.e. if $E_i(x) \neq 0$ and $E_j(x) = 0$ (where $j \neq i$), then $Spr(E(x)) = 1$.
    \item If two explanations have the same sparsity measure for the non-zero part of the attribution but one has more zero attributions, the longer one is more sparse. In other words, if $Spr(E^{\neq 0}(x)) = Spr(E^{\neq 0}(x'))$ and $|E^{=0}(x)| >|E^{=0}(x')| $, where $E^{\neq 0}$ and $E^{=0}$ means the non-zero and zero portion of the attribution, respectively, then $Spr(E(x)) > Spr(E(x'))$.
\end{enumerate}

Assuming that an attribution, $E(x)$ is min-max normalized such that $0 \leq E_i(x) \leq 1$\footnote{If attribution is the same over all time steps, the min-max normalization outputs a sequence of ones.}, we define the sparsity metric as follows:

\begin{equation}
\label{eq-loc}
    Spr(E(x)) = \left( \frac{\sum_{i=1}^{n} (1-E_i(x))}{n-1} \right) ^\beta 
\end{equation}
where $\beta>1$ is a constant hyper-parameter to spread out the metric more uniformly.
Empirically, for majority of datasets/XAIs, the attribution is relatively sparse and it is rarely uniformly distributed. Consequently, in most cases, all methods give a value between 0.90 and 1 which makes it hard to compare one another. Hence, choosing $\beta>1$ allows us to better spread out the values and compare XAI Methods. It is easy to show that it satisfies all above properties.

\begin{table*} [h]
\scriptsize
  \centering
  \begin{tabular}{l|l|l|lllllllll}
    \toprule
    Dataset & Domain & Classifier & \multicolumn{9}{c}{XAI Robustness ($\times 10^{-4}$)} \\
    \cmidrule(r){4-12}
     &  & Robustness & DeepLIFT & GradientSHAP & G. Backprop & I $\times$ G & IG & Kernel SHAP & LIME & Occlusion & Saliency\\
    \midrule
    \parbox[t]{1mm}{\multirow{5}{*}{\rotatebox[origin=c]{90}{AudioMNIST}}} 
    & Time       & 73.7 & 3.7 & 13.5 & 4.9 & 5.5 & 3.4 & 14.4 & 6.3 & 4.0 & 24.3 \\
    & Freq.      & 85.5 & 7.0 & 9.4 & 2.0 & 8.3 & 3.0 & \textbf{10.6} & 6.8 & 10.0 & 11.6 \\
    & Time/Freq. & \textbf{18.9} & \textbf{0.6} & \textbf{5.7} & \textbf{0.3} & \textbf{1.1} & \textbf{0.5} & 11.3 & \textbf{3.1} & \textbf{0.7} & \textbf{5.5} \\
    & Difference & 46.8 & 4.4 & 13.9 & 16.0 & 4.6 & 5.3 & 14.4 & 6.9 & 3.3 & 37.4 \\
    & Min Zero   & 78.3 & 18.3 & 23.0 & 5.0 & 24.5 & 16.1 & 23.8 & 6.7 & 0.8 & 24.6 \\
    
    \midrule
    \parbox[t]{1mm}{\multirow{5}{*}{\rotatebox[origin=c]{90}{FordA}}} 
    & Time       & \textbf{2.5} & 12.6 & 81.6 & 13.5 & 19.3 & 13.2 & 108.8 & 51.7 & 3.6 & 27.0 \\
    & Freq.      & 2.6 & \textbf{1.1} & 34.8 & \textbf{0.8} & \textbf{1.9} & \textbf{1.0} & 79.1 & 39.5 & 2.2 & \textbf{2.6} \\
    & Time/Freq. & 4.2 & 1.7 & \textbf{24.7} & 1.8 & 3.0 & 1.1 & \textbf{57.6} & \textbf{23.5} & 1.0 & 3.8 \\
    & Difference & 19.7 & 20.2 & 92.3 & 38.3 & 44.2 & 22.6 & 66.7 & 56.2 & 4.0 & 96.6 \\
    & Min Zero   & 2.6 & 17.9 & 81.9 & 13.7 & 27.0 & 20.8 & 115.0 & 53.2 & \textbf{0.2} & 27.2 \\
    
    \midrule
    \parbox[t]{1mm}{\multirow{4}{*}{\rotatebox[origin=c]{90}{GunPoint}}} 
    & Time        & 15.2 & 129.5 & 174.9 & 43.7 & 157.8 & 160.2 & 201.9 & 177.2 & 17.9 & 152.7 \\
    & Freq.       & 7.0 & 6.8 & \textbf{61.3} & \textbf{2.0} & 23.1 & 13.6 & \textbf{196.5} & \textbf{163.2} & \textbf{5.9} & \textbf{26.8} \\
    & Difference  & \textbf{3.6} & \textbf{5.6} & 175.5 & 12.2 & \textbf{19.3} & \textbf{6.8} & 217.6 & 178.1 & 9.5 & 43.3 \\
    & Min Zero    & 16.0 & 55.7 & 171.2 & 43.1 & 93.6 & 77.3 & 207.1 & 175.2 & 12.3 & 152.2 \\
    
    
    \midrule
    \parbox[t]{1mm}{\multirow{4}{*}{\rotatebox[origin=c]{90}{ECG5000}}} 
    & Time       & \textbf{4.2} & 35.5 & 160.5 & 44.8 & 72.1 & 35.7 & 237.5 & 188.4 & 22.6 & 94.0 \\
    & Freq.      & 4.7 & 8.2 & 99.5 & \textbf{5.3} & 24.2 & 10.1 & \textbf{201.0} & 186.5 & 20.2 & \textbf{34.5} \\
    & Difference & 6.0 & \textbf{2.7} & \textbf{63.4} & 31.0 & \textbf{9.2} & \textbf{10.0} & 214.3 & \textbf{185.6} & \textbf{4.8} & 60.3 \\
    & Min Zero   & 5.0 & 69.8 & 188.6 & 51.4 & 100.8 & 76.9 & 241.4 & 191.1 & 4.9 & 102.8 \\
    
    \midrule
    \parbox[t]{1mm}{\multirow{5}{*}{\rotatebox[origin=c]{90}{LargeKit.App.}}} 
    & Time       & 5.1 & 1.7 & 30.1 & 2.5 & 2.9 & 1.5 & 87.3 & 36.2 & 3.3 & 13.8 \\
    & Freq.      & 22.4 & 14.7 & 31.3 & 7.5 & 19.7 & 5.7 & 56.8 & 37.3 & 3.2 & 24.5 \\
    & Time/Freq. & \textbf{3.0} & \textbf{1.2} & \textbf{16.0} & \textbf{0.9} & \textbf{2.2} & \textbf{1.2} & \textbf{45.6} & \textbf{13.1} & \textbf{0.6} & \textbf{2.3} \\
    & Difference & 157.7 & 6.4 & 26.1 & 66.9 & 5.0 & 6.1 & 92.1 & 39.9 & 1.2 & 81.6 \\
    & Min Zero   & 5.0 & 1.4 & 30.3 & 2.5 & 2.8 & 1.5 & 54.4 & 35.8 & 2.6 & 13.8 \\
    
    
    \bottomrule
  \end{tabular}
    \caption{Robustness (ResNet)}
  \label{tbl-rbt-resnet}
\end{table*}

\begin{table*}[h]
\scriptsize

  \centering
  \begin{tabular}{l|l|l|lllllllll}
    \toprule
    Dataset & Domain & Classifier & \multicolumn{9}{c}{XAI Robustness ($\times 10^{-4}$)} \\
    \cmidrule(r){4-12}
     &  & Robustness & DeepLIFT & GradientSHAP & G. Backprop & I $\times$ G & IG & Kernel SHAP & LIME & Occlusion & Saliency\\
    \midrule
    \parbox[t]{1mm}{\multirow{5}{*}{\rotatebox[origin=c]{90}{AudioMNIST}}} 
    & Time & 37.9 & 2.0 & 11.4 & 2.7 & 3.6 & 2.6 & 14.4 & 6.3 & 1.5 & 23.7 \\
    & Freq. & \textbf{12.9} & 4.1 & 8.5 & 0.9 & 5.8 & 1.9 & 13.1 & 6.8 & 4.1 & 8.2 \\
    & Time/Freq. & 17.9 & \textbf{0.5} & \textbf{4.9} & \textbf{0.3} & \textbf{0.5} & \textbf{0.3} & 12.6 & \textbf{3.1} & \textbf{0.3} & \textbf{3.3} \\
    & Difference & 99.3 & 1.7 & 9.3 & 40.9 & 2.6 & 1.4 & \textbf{11.8} & 7.0 & 6.1 & 32.6 \\
    & Min Zero & 40.4 & 23.5 & 23.8 & 2.8 & 23.4 & 17.7 & 23.5 & 6.5 & 0.4 & 23.9 \\
    \midrule
    \parbox[t]{1mm}{\multirow{5}{*}{\rotatebox[origin=c]{90}{FordA}}} 
    & Time & 1.7 & 10.2 & 81.7 & 14.7 & 15.5 & 13.2 & 93.6 & 52.2 & 3.3 & 20.6 \\
    & Freq. & 2.2 & \textbf{0.9} & 37.3 & \textbf{0.9} & \textbf{2.5} & \textbf{1.0} & 78.2 & 46.2 & 2.2 & 3.5 \\
    & Time/Freq. & 7.5 & 1.5 & \textbf{23.3} & 3.9 & 2.9 & 1.2 & \textbf{55.8} & \textbf{22.8} & 0.9 & \textbf{3.2} \\
    & Difference & 13.2 & 15.2 & 82.9 & 82.9 & 45.9 & 26.1 & 78.9 & 57.0 & 4.1 & 97.2 \\
    & Min Zero & \textbf{1.9} & 12.1 & 81.4 & 14.9 & 18.3 & 16.5 & 118.0 & 53.7 & \textbf{0.2} & 20.8 \\
    \midrule
    \parbox[t]{1mm}{\multirow{4}{*}{\rotatebox[origin=c]{90}{GunPoint}}} 
    & Time & 6.9 & 53.7 & 149.3 & 54.9 & 133.9 & 117.3 & 245.5 & 180.0 & 12.4 & 124.9 \\
    & Freq. & 9.9 & \textbf{3.0} & \textbf{63.3} & \textbf{1.1} & \textbf{5.6} & \textbf{4.2} & \textbf{194.6} & \textbf{113.1} & 5.7 & \textbf{6.2} \\
    & Difference & \textbf{5.8} & 5.5 & 146.8 & 5.8 & 5.7 & 7.5 & 207.8 & 177.5 & \textbf{5.6} & 13.6 \\
    & Min Zero & 7.1 & 70.8 & 154.3 & 54.5 & 89.8 & 78.4 & 245.2 & 177.1 & 7.6 & 124.1 \\
    \midrule
    \parbox[t]{1mm}{\multirow{4}{*}{\rotatebox[origin=c]{90}{ECG5000}}} 
    & Time & \textbf{0.8} & 29.4 & 134.8 & 20.4 & 29.4 & 17.6 & \textbf{169.5} & 188.6 & 4.1 & 60.3 \\
    & Freq. & 1.8 & 17.2 & 92.0 & \textbf{2.8} & 9.0 & 3.9 & 189.9 & \textbf{172.6} & 7.0 & \textbf{13.5} \\
    & Difference & 3.6 & \textbf{2.2} & \textbf{84.2} & 5.6 & \textbf{5.2} & \textbf{2.0} & 213.3 & 182.7 & 8.3 & 22.1 \\
    & Min Zero & 1.0 & 62.7 & 162.0 & 23.3 & 60.3 & 37.0 & 218.6 & 190.0 & \textbf{3.8} & 66.9 \\
    \midrule
    \parbox[t]{1mm}{\multirow{5}{*}{\rotatebox[origin=c]{90}{LargeKit.App.}}} 
    & Time & 2.7 & 10.0 & 20.7 & 5.8 & 5.6 & 5.0 & 54.2 & 37.2 & 0.7 & 20.4 \\
    & Freq. & 5.4 & 8.1 & 25.6 & 6.4 & 13.2 & 5.3 & 58.1 & 37.2 & 4.5 & 14.1 \\
    & Time/Freq. & \textbf{1.7} & \textbf{1.6} & \textbf{15.2} & \textbf{3.8} & \textbf{1.2} & \textbf{0.6} & \textbf{44.0} & \textbf{14.7} & 0.6 & \textbf{1.5} \\
    & Difference & 60.8 & 6.9 & 26.1 & 80.9 & 4.9 & 4.9 & 84.3 & 38.9 & 3.0 & 78.0 \\
    & Min Zero & 2.6 & 10.1 & 20.8 & 5.8 & 5.5 & 4.8 & 55.2 & 37.1 & \textbf{0.5} & 20.4 \\
    \bottomrule
  \end{tabular}
  \caption{Robustness (InceptionTime)}
  \label{tbl-rbt-inceptiontime}
\end{table*}

\subsection{Sparsity on Different Spaces}
In Table \ref{tbl-resnet} and \ref{tbl-inceptiontime}, we report sparsity along the faithfulness success percentage to roll out sparse but non-faithful explanations. Here, the first number is the success percentage and the number in the parenthesis is the sparsity. We exclude the sparsity for entries with lower than $50\%$ success percentage. For most datasets, LIME only picks a few input features (regardless of the input space) and produces a highly sparse explanation. However, most explanations turn out to be irrelevant in terms of faithfulness. Note that we use default hyper-parameters when possible for all XAI methods. It is possible to make some methods better by tuning the parameters. 
However, such a hyper-parameter search is not only beyond the scope of this paper, but also unnecessary for the purpose of this paper. We emphasize that our goal is to show that the same XAI method can be used on different explanation spaces and generate different quality explanation.

As shown in Table \ref{tbl-resnet}, XAI methods may disagree on the space in which they produce the sparsest explanation. Nevertheless, the sparsity pattern aligns with our intuition when we proposed explanation spaces. For example, AudioMNIST, containing audio signals, is the most sparse in the time/frequency domain in 6 out of 8 faithful methods and the two others are frequency spaces. FordA, a dataset of engine noises, is most sparse in frequency domain in 6 out of 8 faithful methods, as it was shown in Figure \ref{fig:FrodA-time} and \ref{fig:FrodA-freq}. 
As we speculated, both ECGFiveDays and ECG5000 are sparsest in difference space for most XAI methods\footnote{Note that we exclude experiment with time/frequency domain in datasets where the number of time steps is small because it was impossible to get a better resolution than the frequency domain.}. The last three datasets, LargeKitchenAppliance, ElectricDevices, and Earthquakes, are examples of datasets where the original zero value (before normalization) corresponds to the lack of feature and, hence, 
min zero space is the most sparse space across most XAI methods. Overall, the only exception was the GunPoint where we initially hypothesized that the difference space is the most appropriate due to the nature of the dataset, containing hand movement. However, by comparing time and frequency domains (as depicted in Figure \ref{fig:gun-time} and \ref{fig:gun-freq}), it is evident that only the first few frequency components are non-negligible and others are negligible. Hence, we speculate that for datasets containing only low frequencies, sparsity metric is smaller for frequency domain. Overall, we have observed similar pattern in InceptionTime results shown in Table \ref{tbl-inceptiontime}.



\subsection{Robustness on Different Spaces}

The robustness results of ResNet model is shown in Table \ref{tbl-rbt-resnet}. Due to the lack of space, we show only one dataset for each dataset type. 
Unlike sparsity metric, there is no strong correlation between dataset types and robustness. However, there are a few observations worth considering: First, classifier robustness does not necessarily match the XAI robustness. For example, the time domain is the most robust space in ECG5000 while the XAI robustness on time domain is among the worst. Interestingly, whenever time/frequency is provided for a dataset, XAI methods on this space are among the most robust ones (20 out of 27). Unlike the classifier robustness where the min zero space and time domain perform similarly, their XAI robustness can be vastly different due to the way they handle the baseline. Overall, no clear indication can be given on the best explanation space purely based on the robustness. Results of InceptionTime reveal similar pattern, shown in Table \ref{tbl-rbt-inceptiontime}.

\begin{figure}[]
    \centering
    \subfloat[Original Time Domain]{%
        \includegraphics[width=0.8\linewidth, height=0.12\textwidth]{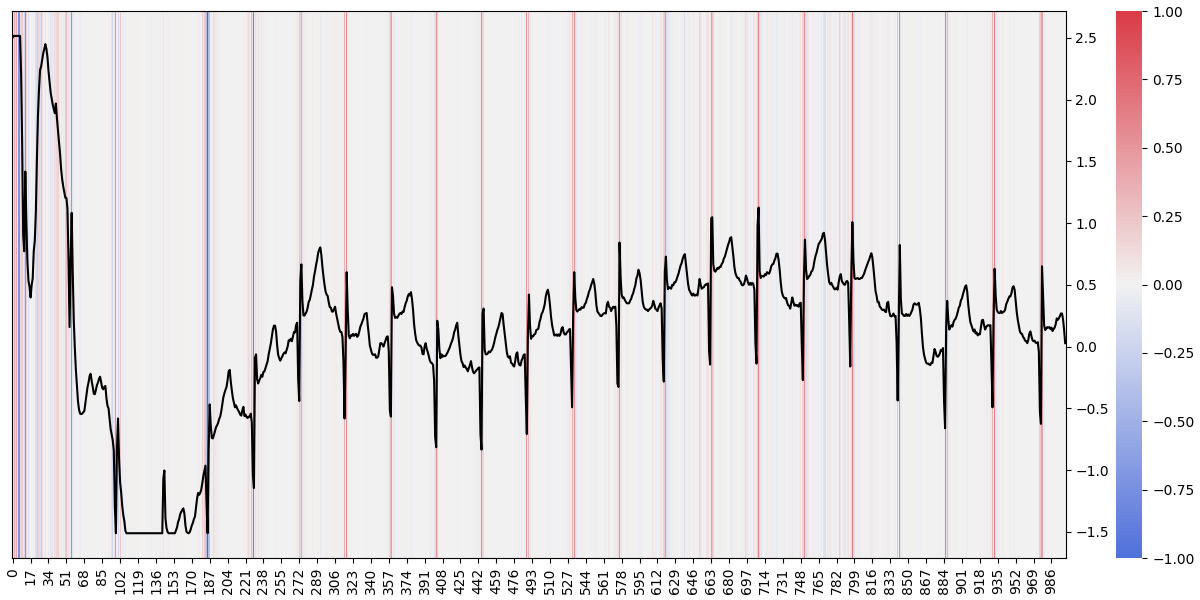}
        \label{fig:mimic0}}
    \\
    \subfloat[Decomposed component 1]{%
        \includegraphics[width=0.8\linewidth, height=0.12\textwidth]{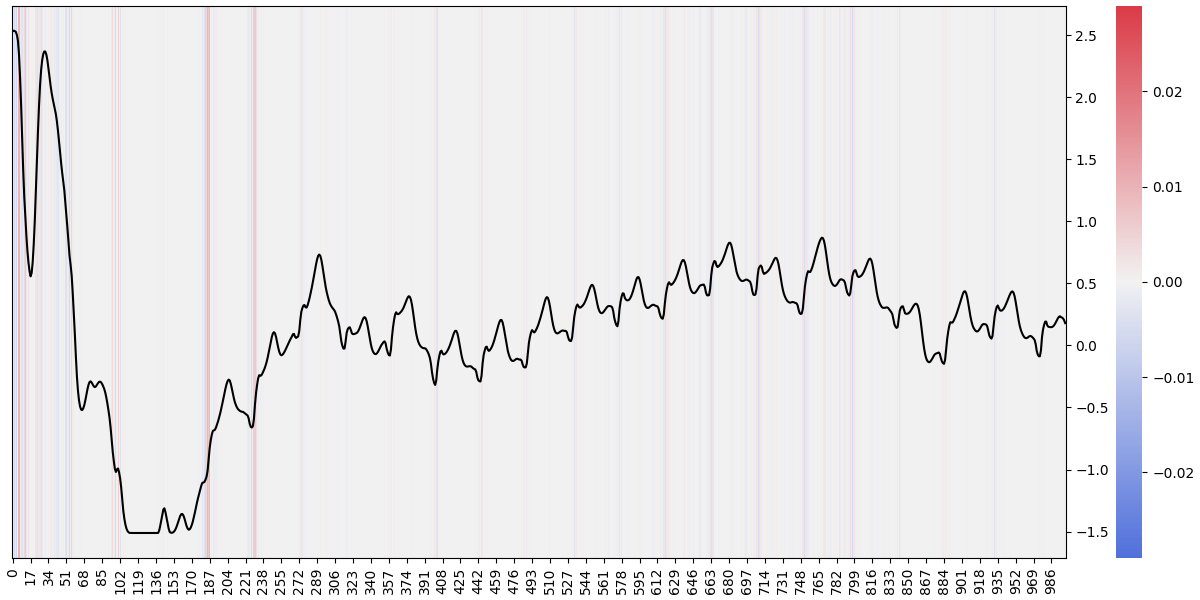}
        \label{fig:mimic1}}
    \\
    \subfloat[Decomposed component 2]{%
        \includegraphics[width=0.8\linewidth, height=0.12\textwidth]{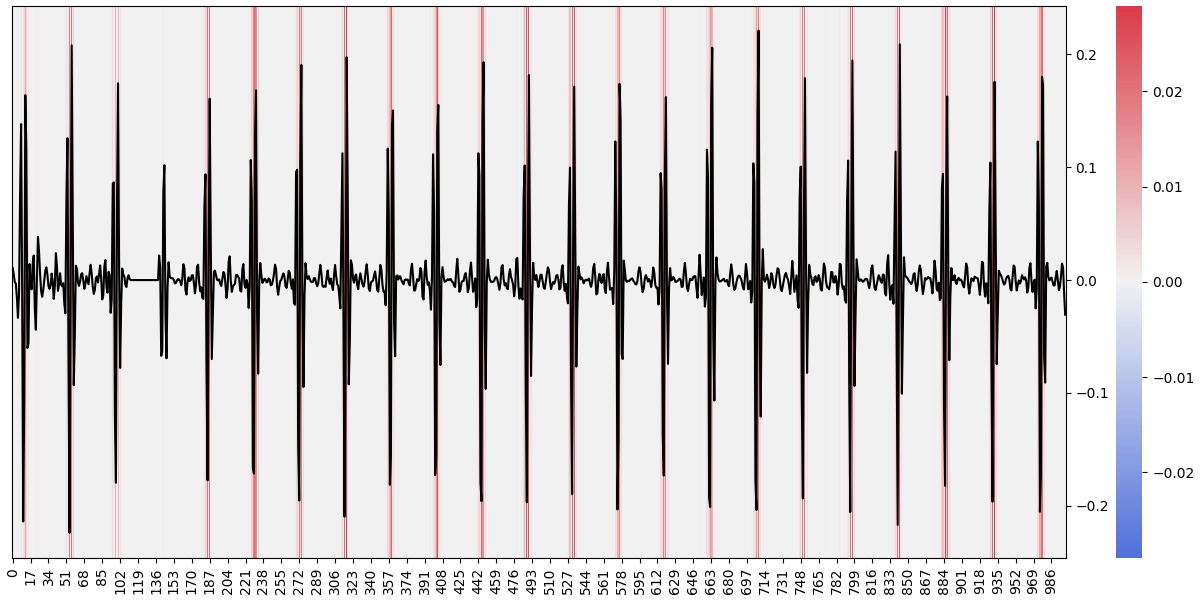}
        \label{fig:mimic2}}
    \\
    \subfloat[Decomposed component 3]{%
        \includegraphics[width=0.8\linewidth, height=0.12\textwidth]{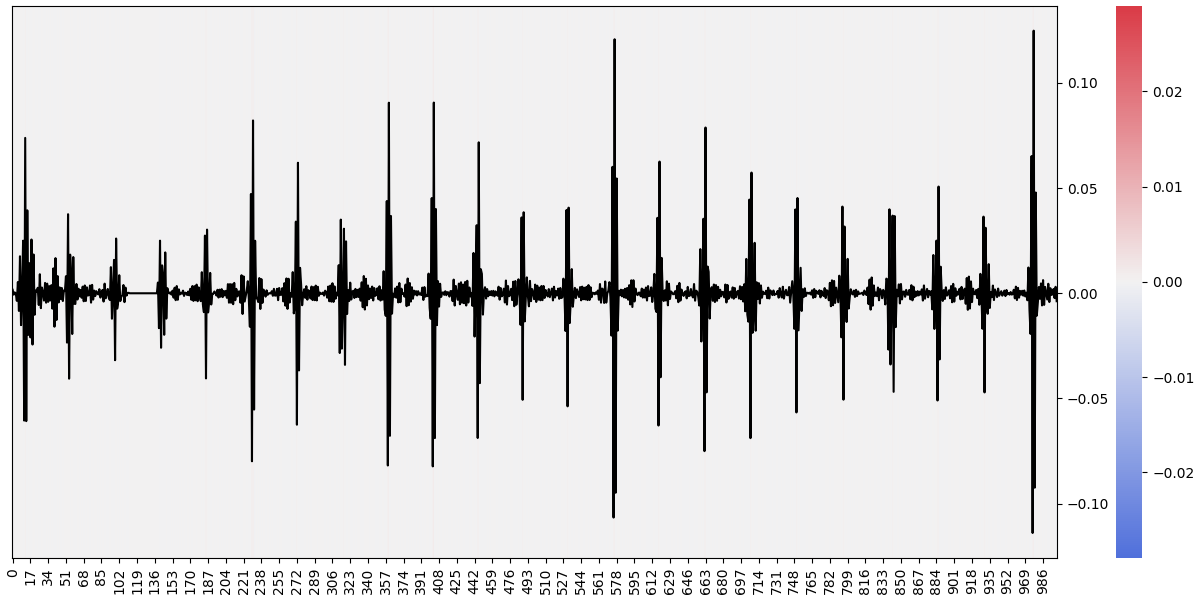}
        \label{fig:mimic3}}
    \caption{DeepLIFT attribution method on MIMIC Performance dataset.}
    \label{fig:mimic}
\end{figure}

\section{Decomposition Space}
To decompose a time series instance, we use singular-spectrum analysis (SSA) implementation provided in Kaggle\footnote{ https://www.kaggle.com/code/jdarcy/introducing-ssa-for-time-series-decomposition}. Unfortunately, none of the datasets in the UCR repository is suitable for decomposition since they are often short with no trends or seasonality. To see the usefulness of the decomposition space, we use the MIMIC performance dataset \cite{mimic} and US hourly climate dataset \cite{ushourly}.

\begin{figure}[t]
    \centering
    \subfloat[Original Time Domain]{%
        \includegraphics[width=0.8\linewidth, height=0.12\textwidth]{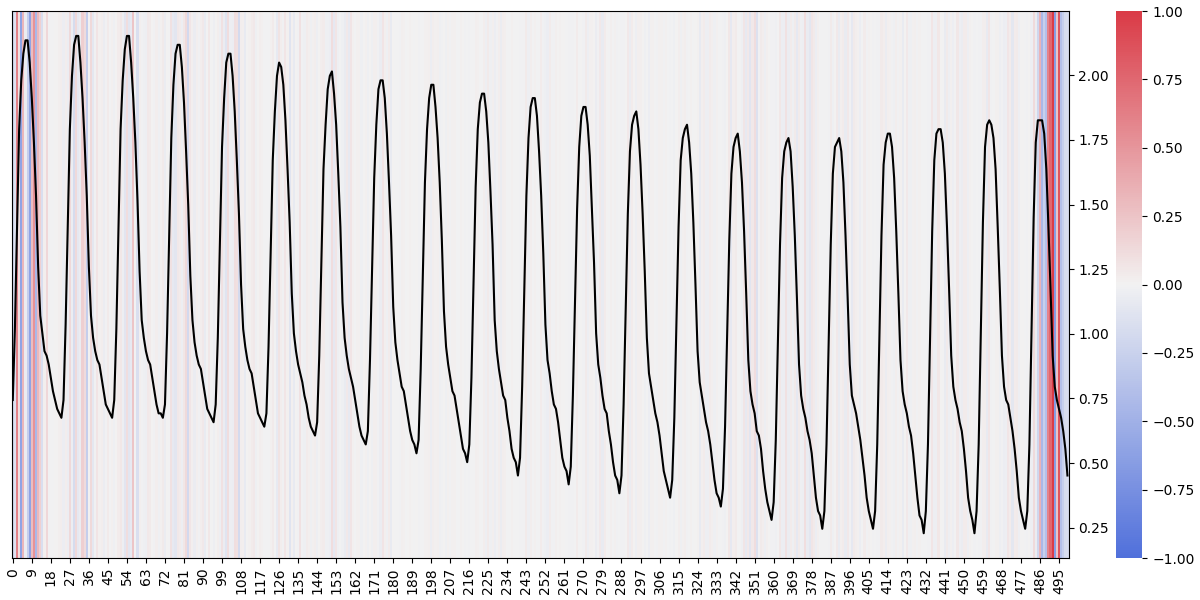}
        \label{fig:us0}}
    \\
    \subfloat[Decomposed component 1]{%
        \includegraphics[width=0.8\linewidth, height=0.12\textwidth]{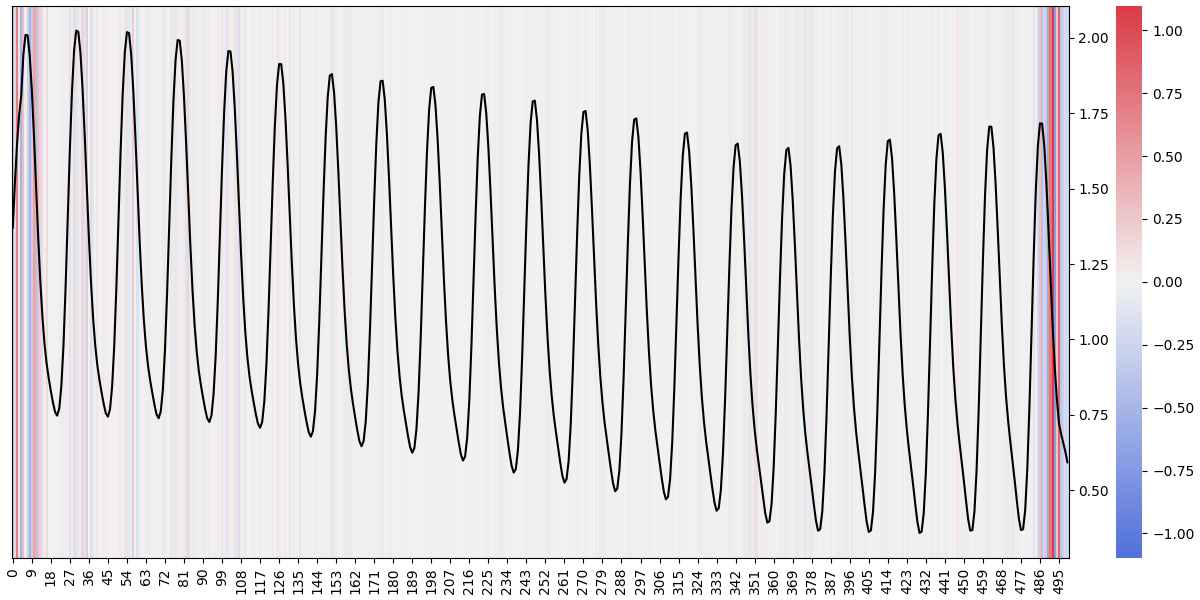}
        \label{fig:us1}}
    \\
    \subfloat[Decomposed component 2]{%
        \includegraphics[width=0.8\linewidth, height=0.12\textwidth]{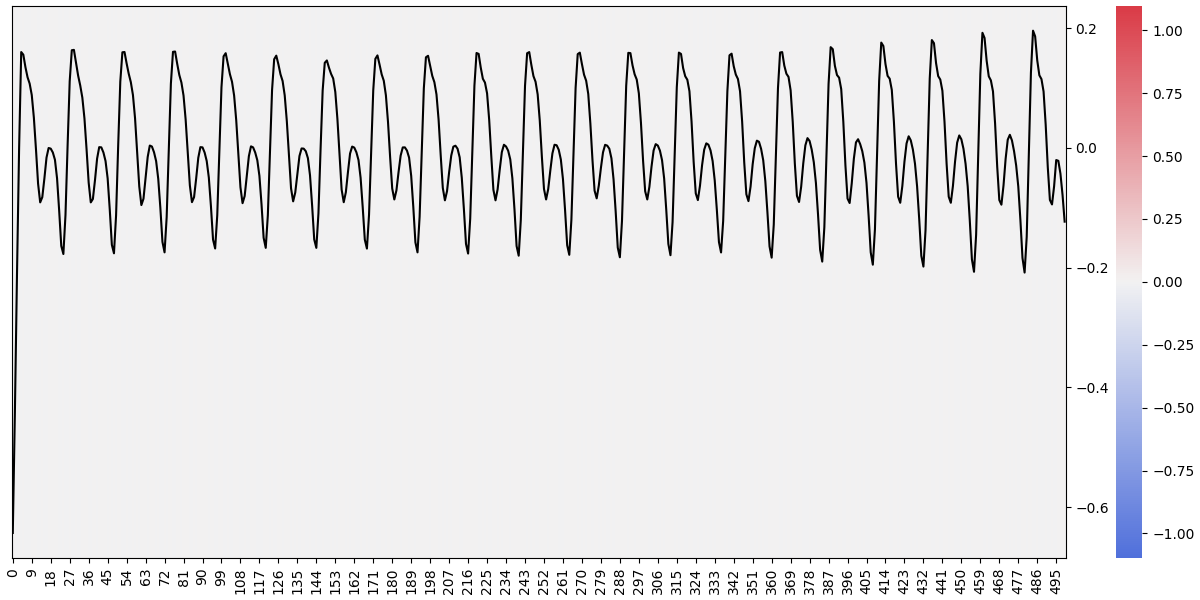}
        \label{fig:us2}}
    \\
    \subfloat[Decomposed component 3]{%
        \includegraphics[width=0.8\linewidth, height=0.12\textwidth]{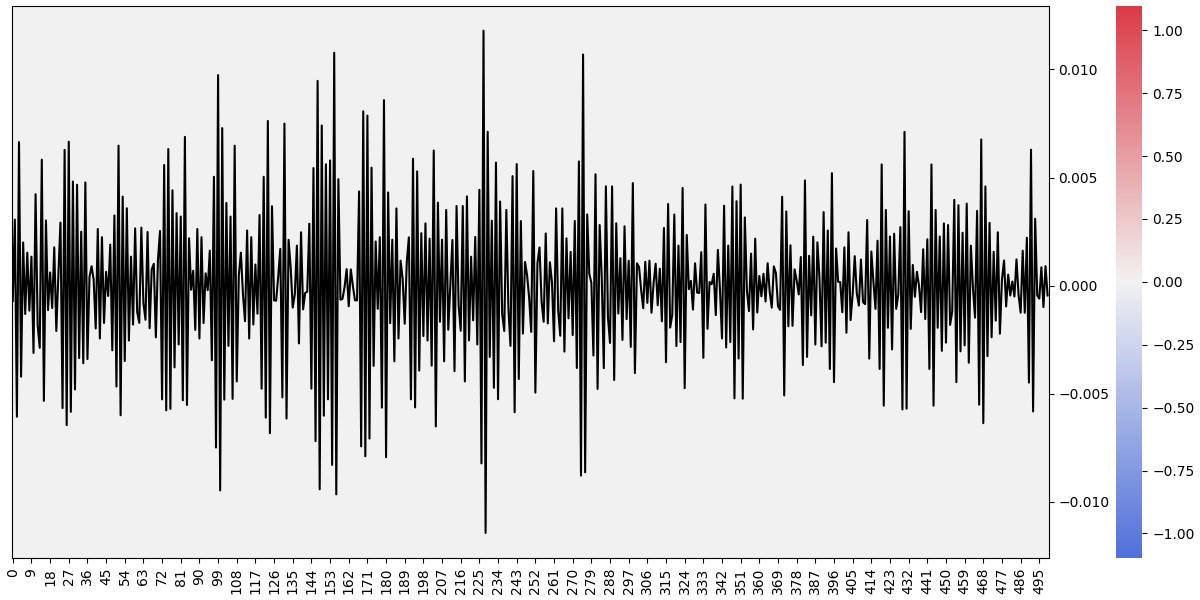}
        \label{fig:us3}}
    \caption{Input $\times$ Gradient attribution method on US hourly dataset.}
    \label{fig:us}
\end{figure}

The MIMIC performance dataset contains PPG, ECG, and respiration time series and the task is to identify adults versus neonates. Here, we only use the ECG time series. We subsample each subjects' ECG records such that each sample has a fixed length of 1000 time steps. In each sample, there are on average 25 heartbeats. Hence, the samples are long enough to capture trends or cycles. US hourly climate dataset is a public dataset provided by the National Centers of Environmental Information. It contains hourly temperature and wind speed recorded at thousands of stations across united state. Here, we use the Los Angeles airport station (USW00023174) data. We use the first $90\%$ of the time series as a training set and the last $10\%$ for validation. The length of each sample is 500 which almost covers three weeks of data. The task is to predict whether the next time step of the temperature increases or decreases. We train the ResNet50 model on the original times series (non-decomposed ones). The validation accuracy on the MIMIC and US hourly datasets was $97.93\%$ and $96.37\%$, respectively.

Figure \ref{fig:mimic} shows the attribution of the original time series versus the decomposition space components for the MIMIC performance dataset. Decomposition component 1 mostly captures the trend plus a small component of the QRS heartbeat complex. From the decomposed component 2 that has the most cyclic component of the QRS complex, it is easy to see that the model relies mainly on this cyclic component rather than the overall trend or the remaining part (the last component). This interpretation is not clearly accessible by looking at the original time series because the components are all mixed together.

Figure \ref{fig:us} demonstrates the attribution for the US hourly temperature dataset. Looking at the attribution in the original time domain, it is clear that the model relies mainly on the last few time points, which is reasonable for forecasting the next time step. However, it is difficult to understand what component has the most effect. In time series analysis, decomposition is often used to separate trends and cycles from the remaining part. Our approach allows different components to be studied separately. In this example, the decomposition space clearly shows that the main cycle, component 1, with largest amplitude is mostly used by the model, while the remaining two components have little effect on the model performance.

\section{Discussion}

In this paper, we bridge the gap between application-blind approaches in computer science community and the application-specific approaches used by domain experts. Our method allows computer scientists to continue proposing application-blind XAI methods without worrying about the explanation space. Interdisciplinary researchers can now focus on proposing new mappings to project known spaces to desired, yet ignored, spaces, to facilitate the usage of existing XAI. Finally, our framework allows domain experts to easily explore other spaces and investigate if there exists a better space for explanation.
This can also help practitioners to investigate what features a model uses. As shown in \cite{vielhaben2024explainable}, a model that exploits average value of the entire time series is hard to catch in time domain. However, it can be easily identified by checking the first frequency component of the frequency space.

\bibliographystyle{IEEEtran}
\bibliography{IEEEabrv,mybibfile}

\section{Appendices}

\subsection{Explanation Space Implementation}
The representation space function, $F(x)$, is used outside a neural network and it can be construed as a pre-processing stage (with respect to the explanation generation stage at inference time, not with respect to training stage). Hence, the target domain data, $z=F(x)$, can be obtained by implementing the transformation in any arbitrary way. However, $F^{-1}(z)$ function is embedded into the $M'$ neural network. Hence, for compatibility with gradient-based methods, we either need to use the existing functions in deep learning packages or implement the function using existing layers. Fortunately, most packages, like PyTorch, has the implementation of FFT, STFT (through torchaudio.transforms.Spectrogram), and their inverse. We implement inverse mapping function for min zero space and difference space using a single fully connected layer with a pre-defined weight matrix, as follows:

\textbf{Min Zero Space Implementation: } The inverse mapping of the min zero space can be written as follows:
\begin{multline}
    F^{-1}_{min\_zero} (\{x_1 - x_{min}, ..., x_N -x_{min}, x_{min}\}) \rightarrow \\
    \{x_1, x_2, ..., x_N\}
\end{multline}

In terms of a fully connected layer, it can be implemented as a layer that takes a vector of size $N+1$ as input and output a vector of size $N$, where the $(N+1)^{th}$ element is added to all other elements in the input vector and then it is dropped from the output. It can be easily shown that an Identity matrix of size $N \times N$ concatenated to an all-one matrix of size $1 \times N$ can obtain the reverse mapping as a linear layer as follows:

\begin{multline} 
     \begin{bmatrix}
         x_1 - x_{min}\\
         x_2 - x_{min}\\
         ... \\
         x_{N-1} - x_{min} \\
         x_N - x_{min} \\
         x_{min} \\
     \end{bmatrix}_{N+1 \times 1}^T
     \begin{bmatrix}
        1 & 0 & ... & 0 & 0 \\
        0 & 1 & ... & 0 & 0 \\
        ... & ... & ... & ... & ... \\
        0 & 0 & ... & 1 & 0 \\
        0 & 0 & ... & 0 & 1 \\
        1 & 1 & ... & 1 & 1 \\
        \end{bmatrix}_{N+1 \times N} \\
        =
        \begin{bmatrix}
             x_1\\
             x_2\\
             ... \\
             x_{N-1}\\
             x_N\\
         \end{bmatrix}_{N \times 1}^T
\end{multline}

\textbf{Difference Space Implementation: } The inverse mapping of the difference space can be written as follows:

\begin{multline}
    F^{-1}_{diff} (\{x_1, x_2-x_1, x_3-x_2, ..., x_N-x_{N-1}\}) \rightarrow \\
    \{x_1, x_2, ..., x_N\}
\end{multline}

It can be easily shown that the $i^{th}$ elements of the output can be obtained by adding all previous input elements up to $i^{th}$ element. As a result, all $1$ to $i-1$ elements are cancel out and only the $x_i$ remains which constructs the inverse operation. In linear layer with matrix multiplication operation, it can be easily implemented by a weight matrix of upper triangular form where every non-zero element is exactly one, as follow:

\begin{multline} 
     \begin{bmatrix}
         x_1\\
         x_2 - x_1\\
         ... \\
         x_{N-1} - x_{N-2} \\
         x_{N} - x_{N-1} \\
     \end{bmatrix}_{N \times 1}^T
     \begin{bmatrix}
        1 & 1 & ... & 1 & 1 \\
        0 & 1 & ... & 1 & 1 \\
        ... & ... & ... & ... & ... \\
        0 & 0 & ... & 1 & 1 \\
        0 & 0 & ... & 0 & 1 \\
        \end{bmatrix}_{N \times N} \\
        =
        \begin{bmatrix}
             x_1\\
             x_2\\
             ... \\
             x_{N-1}\\
             x_N\\
         \end{bmatrix}_{N \times 1}^T
\end{multline}

\end{document}